\documentclass{article}

\PassOptionsToPackage{numbers, compress}{natbib}
\usepackage[final]{neurips_2022}

\usepackage[utf8]{inputenc} % allow utf-8 input
\usepackage[T1]{fontenc}    % use 8-bit T1 fonts
\usepackage{hyperref}       % hyperlinks
\usepackage{url}            % simple URL typesetting
\usepackage{booktabs}       % professional-quality tables
\usepackage{amsfonts}       % blackboard math symbols
\usepackage{nicefrac}       % compact symbols for 1/2, etc.
\usepackage{microtype}      % microtypography
\usepackage{xcolor}         % colors
\usepackage[pdftex]{graphicx}
\usepackage[symbol]{footmisc}
\usepackage{amsmath}
\usepackage{chngpage}
\usepackage{bm}

\usepackage{xcolor}
\usepackage[export]{adjustbox}[2011/08/13]
\usepackage[acronym]{glossaries}
\usepackage{amsthm}
\newtheorem{thm}{Theorem}[]
\newtheorem{lem}{Lemma}[]

\newcommand{\dd}{\,\mathrm{d}}
\newcommand{\R}{\mathbb{R}}

\renewcommand{\H}{\mathcal{H}}

\usepackage{comment}
\usepackage{algorithm2e}
\usepackage{algpseudocode}

\usepackage{comment}
\usepackage{pifont}

\title{Fast Bayesian Inference with Batch Bayesian Quadrature via Kernel Recombination}

\author{%
  Masaki Adachi\thanks{Equal contribution} \\
  \small{Machine Learning Research Group, University of Oxford}\\
  \small{Toyota Motor Corporation}\\
  \small{\texttt{masaki@robots.ox.ac.uk}}\\
  \And
  Satoshi Hayakawa$^{*}$, Harald Oberhauser\\
  \small{Mathematical Institute, University of Oxford}\\
  \small{\texttt{\{hayakawa,oberhauser\}@maths.ox.ac.uk}}\\
  \And
  Martin Jørgensen, Michael A. Osborne\\
  \small{Machine Learning Research Group, University of Oxford}\\
  \small{\texttt{\{martinj, mosb\}@robots.ox.ac.uk}}
}

\begin{document}

\maketitle

\begin{abstract}
Calculation of Bayesian posteriors and model evidences typically requires numerical integration.
Bayesian quadrature (BQ), a surrogate-model-based approach to numerical integration, is capable of superb sample efficiency, but its lack of parallelisation has hindered its practical applications.
In this work, we propose a parallelised (batch) BQ method, employing techniques from kernel quadrature, that possesses an empirically exponential convergence rate.
Additionally, just as with Nested Sampling, our method permits simultaneous inference of both posteriors and model evidence.
Samples from our BQ surrogate model are re-selected to give a sparse set of samples, via a kernel recombination algorithm, requiring negligible additional time to increase the batch size.
Empirically, we find that our approach significantly outperforms the sampling efficiency of both state-of-the-art BQ techniques and Nested Sampling in various real-world datasets, including lithium-ion battery analytics.\footnote{Code: \url{https://github.com/ma921/BASQ}}\looseness=-1
\end{abstract}

\newacronym{mc}{MC}{Monte Carlo}
\newacronym{smc}{SMC}{Sequential Monte Carlo}
\newacronym{ns}{NS}{nested sampling}
\newacronym{bq}{BQ}{Bayesian quadrature}
\newacronym{gp}{GP}{Gaussian process}
\newacronym{wsabi}{WSABI}{Warped sequential active Bayesian integration}
\newacronym{wsabil}{WSABI-L}{WSABI with linearisation approximation}
\newacronym{kq}{KQ}{kernel quadrature}
\newacronym{rchq}{RCHQ}{\textit{random convex hull quadrature}}
\newacronym{basq}{BASQ}{\textit{Bayesian alternately subsampled quadrature}}
\newacronym{mae}{MAE}{mean absolute error}
\newacronym{rkhs}{RKHS}{reproducing kernel Hilbert space}
\newacronym{fwbq}{FWBQ}{Frank-Wolfe Bayesian quadrature}
\newacronym{mle}{MLE}{maximum likelihood estimation}
\newacronym{map}{MAP}{maximum a posteriori probability}
\newacronym{igb}{IGB}{initial guess believer}
\newacronym{ub}{UB}{uncertainty believer}
\newacronym{ivr}{IVR}{integral variance reduction}
\newacronym{kl}{KL}{Kullback-Leibler}
\newacronym{kde}{KDE}{kernel density estimation}
\newacronym{rmse}{RMSE}{root mean squared error}
\newacronym{spme}{SPMe}{SPMe}
\newacronym{lbfgs}{L-BFGS}{L-BFGS}
\newacronym{pfm}{PFM}{phase-field model}
\newacronym{af}{AF}{acquisition function}

\section{Introduction}
%\renewcommand{\thefootnote}{\fnsymbol{footnote}}
%\footnote[1]{Equal Contribution}
Many applications in science, engineering, and economics involve complex simulations to explain the structure and dynamics of the process.
Such models are derived from knowledge of the mechanisms and principles underlying the data-generating process, and are critical for scientific hypothesis-building and testing.
However, dozens of plausible simulators describing the same phenomena often exist, owing to differing assumptions or levels of approximation.
Similar situations can be found in selection of simulator-based control models, selection of machine learning models on large-scale datasets, and in many data assimilation applications \citep{evensen1994sequential}.

In such settings, with multiple competing models, choosing the best model for the dataset is crucial.
Bayesian model evidence gives a clear criteria for such model selection. However, computing model evidence requires integration over the likelihood, which is challenging, particularly when the likelihood is non-closed-form and/or expensive.
The ascertained model is often applied to produce posteriors for prediction and parameter estimation afterwards.
There are many algorithms specialised for the calculation of model evidences or posteriors, although only a limited number of Bayesian inference solvers estimate both model evidence \emph{and} posteriors in one go.
As such, costly computations are often repeated (at least) twice.
Addressing this concern, \gls{ns} \citep{skilling2006nested, higson2019dynamic} was developed to estimate both model evidence and posteriors simultaneously, and has been broadly applied, especially amongst astrophysicists for cosmological model selection \citep{mukherjee2006nested}.
However, \gls{ns} is based on a \gls{mc} sampler, and its slow convergence rate is a practical hindrance.

To aid \gls{ns}, and other approaches, parallel computing is widely applied to improve the speed of wall-clock computation.
Modern computer clusters and graphical processing units enable scientists to query the likelihood in large batches. However, parallelisation can, at best, linearly accelerate \gls{ns}, doing little to counter \gls{ns}'s inherently slow convergence rate as a \gls{mc} sampler.

This paper investigates batch \gls{bq} \citep{ohagan1991bayes} for fast Bayesian inference. \gls{bq} solves the integral as an inference problem, modelling the likelihood function with a probabilistic model (typically a \gls{gp}). Gunter et al. \citep{gunter2014sampling} proposed \gls{wsabi}, which adopts active learning to select samples upon uncertainty over the integrand. \gls{wsabi} showed that \gls{bq} with expensive \gls{gp} calculations could achieve faster convergence in wall time than cheap \gls{mc} samplers. Wagstaff et al. \citep{wagstaff2018batch} introduced batch \gls{wsabi}, achieving even faster calculation via parallel computing and became the fastest \gls{bq} model to date. We improve upon these existing works for a large-scale batch case.

\section{Background}
\paragraph{Vanilla Bayesian quadrature}
While \gls{bq} in general is the method for the integration, the functional approximation nature permits solving the following integral $Z$ and obtaining the surrogate function of posterior $p(x)$ simultaneously in the Bayesian inference context:

\begin{equation}
    p(x) = \frac{\ell_\text{true}(x) \pi(x)}{Z}  = \frac{\ell_\text{true}(x) \pi(x)}{\int \ell_\text{true}(x) \pi(x) \dd x}, \label{eq:BQ-general}
\end{equation}

where both $\ell_\text{true}(x)$ (e.g. a likelihood) and $\pi(x)$ (e.g. a prior) are non-negative, and $x \in \mathbb{R}^d$ is a sample, and is sampled from prior $x \sim \pi(x)$.
\gls{bq} solves the above integral as an inference problem, modelling a likelihood function $\ell(x)$ by a \gls{gp} in order to construct a surrogate model of the expensive true likelihood $\ell_\text{true}(x)$. The surrogate likelihood function $\ell(x)$ is modelled:

\begin{subequations}
\begin{align}
 \ell \,|\, \textbf{y} &\sim \mathcal{GP}(\ell; m_{\textbf{y}}, C_{\textbf{y}}), \\
 m_{\textbf{y}}(x) &= K(x, \textbf{X})K(\textbf{X}, \textbf{X})^{-1} \textbf{y}, \\
 C_{\textbf{y}}(x, x') &= K(x, x') - K(x, \textbf{X})K(\textbf{X}, \textbf{X})^{-1}K(\textbf{X}, x'),
 \end{align}\label{eq:GP}
 \end{subequations}
where $\textbf{X} \in \mathbb{R}^{n \times d}$ is the matrix of observed samples, $\textbf{y} \in \mathbb{R}^n$ is the observed true likelihood values, $K$ is the kernel.
\footnote{
In GP modelling, the GP likelihood function is modelled as $\mathcal{GP}(0, K)$, and \eqref{eq:GP} is the resulting posterior GP. Throughout the paper, we refer to a symmetric positive semi-definite kernel just as a kernel. The notations $\sim$ and $|$ refer to being sampled from and being conditioned, respectively.
}
Due to linearity, the mean and variance of the integrals are simply
\begin{subequations}
\begin{align}
 \mathbb{E}[Z \,|\,\textbf{y}] &= \int m_{\textbf{y}}(x) \pi(x)\dd x, \\
 \mathbb{V}\text{ar}[Z \,|\, \textbf{y}] &= \iint C_{\textbf{y}}(x, x^\prime) \pi(x) \pi(x^\prime) \dd x\dd x^\prime.
 \end{align}\label{eq:BQ}
 \end{subequations}

In particular, \eqref{eq:BQ} becomes analytic when $\pi(x)$ is Gaussian and $K$ is squared exponential kernel, $K(\textbf{X}, x) = v \sqrt{|2\pi \textbf{W}|} \mathcal{N}(\textbf{X}; x, \textbf{W})$, where $v$ is kernel variance and $\textbf{W}$ is the diagonal covariance matrix whose diagonal elements are the lengthscales of each dimension. Since both the mean and variance of the integrals can be calculated analytically, posterior and model evidence can be obtained simultaneously. Note that non-Gaussian prior and kernel are also possible to be chosen for modelling via kernel recombination (see Supplementary). Still, we use this combination throughout this paper for simplicity.

\paragraph{Warped Bayesian quadrature (WSABI)}
\gls{wsabil} adopts the square-root warping \gls{gp} for non-negativity with linearisation approximation of the transform $\tilde \ell \mapsto \ell = \alpha + \frac{1}{2}\tilde \ell^2 $.
\footnote{
$\alpha := 0.8 \times \text{min}(\textbf{y})$. See Supplementary for the details on $\alpha$}
The square-root \gls{gp} is defined as $\tilde \ell \sim \mathcal{GP}(\tilde \ell; \tilde m_{\textbf{y}}, \tilde C_{\textbf{y}})$,
and we have the following linear approximation:
\begin{subequations}
\begin{align}
 \ell \,|\, \textbf{y} &\sim \mathcal{GP}(\ell; m_{\textbf{y}}^{L}, C_{\textbf{y}}^{L}), \\
 m_{\textbf{y}}^{L}(x) &= \alpha + \frac{1}{2} \tilde m_{\textbf{y}}(x)^2, \\
 C_{\textbf{y}}^{L}(x, x') &= \tilde m_{\textbf{y}}(x) \tilde C_{\textbf{y}}(x, x') \tilde m_{\textbf{y}}(x').
 \end{align}\label{eq:WSABI}
\end{subequations}
Gaussianity implies the model evidence $Z$ and posterior $p(x)$ remain analytical (see Supplementary).

%\subsection{Kernel recombination}\label{sec:kerquad}
\paragraph{Kernel quadrature in general}
\gls{kq} is the group of
numerical integration rules
for calculating the integral of function classes that form the \gls{rkhs}.
With a \gls{kq} rule $Q_{\textbf{w},\textbf{X}}$
given by weights $\textbf{w}=(w_i)_{i=1}^n$
and points $\textbf{X} = (x_i)_{i=1}^n$,
we approximate the integral by the weighted sum
\begin{equation}
    Q_{\textbf{w}, \textbf{X}}(h):= \sum^n_{i=1} w_i h(x_i) \approx \int h(x) \pi(x)\dd x,
    \label{eq:kq-general}
\end{equation}
where $h$ is a function of \gls{rkhs} $\H$ associated with the kernel $K$.
We define its worst-case error
by $\text{wce}(Q_{\textbf{w}, \textbf{X}}) :=
\sup_{\lVert h\rVert_{\H}\le 1}\lvert Q_{\textbf{w}, \textbf{X}}(h)- \int h(x)\pi(x)\dd x\rvert$.
Surprisingly, it is shown in \citep{huszar2012optimally} that we have
\begin{equation}
    \mathbb{V}\text{ar}[Z\,|\,\textbf{y}] = \inf_{\textbf{w}}\text{wce}(Q_{\textbf{w}, \textbf{x}})^2.
    \label{eq:bq-kq-equivalence}
\end{equation}
Thus, the point configuration in \gls{kq} with a small worst-case error gives a good way to select points to reduce the integral variance
in Bayesian quadrature.

\paragraph{Random Convex Hull Quadrature (RCHQ)}
Recall from \eqref{eq:kq-general} and \eqref{eq:bq-kq-equivalence}
that we wish to approximate the integral of a function $h$ in
the current \gls{rkhs}.
First, we prepare $n-1$ test functions $\varphi_1,\ldots,\varphi_{n-1}$
based on $M$ sample points using the Nystr{\"o}m approximation of the kernel:
$\varphi_i(x) := u_i^\top K(\textbf{X}_\text{nys}, x)$,
where $u_i\in\R^M$ is the $i$-th eigenvector of $K(\textbf{X}_\text{nys}, \textbf{X}_\text{nys})$.
If we let $\lambda_i$ be the $i$-th eigenvalue of the same matrix,
the following gives a practical approximation \citep{kum12}:
\begin{equation}
    K_0(x, y):=\sum_{i=1}^{n-1}\lambda_i^{-1}\varphi_i(x)\varphi_i(y).
    \label{eq:nystrom}
\end{equation}

Next, we consider extracting a weighted set of $n$ points
$(\textbf{w}_\text{quad}, \textbf{X}_\text{quad})$
from a set of $N$ points
$\textbf{X}_\text{rec}$ with positive weights $\textbf{w}_\text{rec}$.
We do it by the so-called kernel recombination algorithm \citep{lit12,tch15},
so that the measure induced by $(\textbf{w}_\text{quad}, \textbf{X}_\text{quad})$
exactly integrates the above test functions $\varphi_1,\ldots, \varphi_{n-1}$
with respect to the measure given by $(\textbf{w}_\text{rec}, \textbf{X}_\text{rec})$
\citep{hayakawa2021positively}.

In the actual implementation of multidimensional case,
we execute the kernel recombination
not by the algorithm \citep{tch15} with
the best known computational complexity $\mathcal{O}(C_\varphi N + n^3\log(N/n))$ (where $C_\varphi$ is the cost of evaluating $(\varphi_i)_{i=1}^{n-1}$ at a point),
but the one of \citep{lit12} using an LP solver (Gurobi \citep{gurobi2022gurobi} for this time)
with empirically faster computational time.
We also adopt the randomized SVD \citep{hal11} for
the Nystr{\"o}m approximation,
so we have a computational time empirically faster than
$\mathcal{O}(NM + M^2\log n + Mn^2\log(N/n))$ \citep{hayakawa2021positively} in practice.

\section{Related works}
\paragraph{Bayesian inference for intractable likelihood}
Inference with intractable likelihoods is a long-standing problem, and a plethora of methods have been proposed. Most infer posterior and evidence separately, and hence are not our fair competitors, as solving both is more challenging. For posterior inference, \textit{Markov Chain Monte Carlo} \citep{metropolis1953equation, hasting1970monte}, particularly \textit{Hamilton Monte Carlo} \citep{hoffman2014the}, is the gold standard. In a \textit{Likelihood-free inference} context, \gls{kde} with Bayesian optimisation \citep{gutmann2016bayesian} and neural networks \citep{greenberg2019automatic} surrogates are proposed for simulation-based inference \citep{cranmer2020frontier}. In a \textit{Bayesian coreset} context, scalable Bayesian inference \citep{manousakas2020bayesian}, sparse variational inference \citep{campbell2019sparse, campbell2019automated}, active learning \citep{pinsler2019bayesian} have been proposed for large-scale dataset inference. However, all of the above only calculate posteriors, not model evidences.
For evidence inference, \textit{Annealed Importance Sampling} \citep{neal2001annealed}, and \textit{Bridge Sampling} \citep{bennett1976efficient}, \gls{smc} \citep{didelot2011likelihood} are popular, but \emph{only} estimate evidence, not the posterior.

\paragraph{Bayesian quadrature}
The early works on \gls{bq}, which directly replaced the likelihood function with a \gls{gp} \citep{ohagan1991bayes, ohagan1998bayesian, rasmussen2003bayesian}, did not explicitly handle non-negative integrand constraints. Osborne et al. \citep{osborne2012active} introduced logarithmic warped \gls{gp} to handle the non-negativity of the integrand, and introduced active learning for \gls{bq}, a method that selects samples based on where the variance of the integral will be minimised. Gunter et al. \citep{gunter2014sampling} introduced square-root \gls{gp} to make the integrand function closed-form and to speed up the computation. Furthermore, they accelerated active learning by changing the objective from the variance of the integral $\mathbb{V}\text{ar}[Z | \textbf{y}]$ to simply the variance of the integrand $\mathbb{V}\text{ar}[\ell(x)\pi(x)]$. Wagstaff et al. \citep{wagstaff2018batch} introduced the first batch \gls{bq}. Chai et al. \citep{Chai2016improving} generalised warped \gls{gp} for \gls{bq}, and proposed probit-transformation. \gls{bq} has been extended to machine learning tasks (model selection \citep{chai2019automated}, manifold learning \citep{frohlich2021bayesian}, kernel learning \citep{hamid2021marginalising}) with new \gls{af} designed for each purpose. For posterior inference, VBMC \citep{acerbi2018variational} has pioneered that \gls{bq} can infer posterior and evidence in one go via variational inference, and \citep{acerbi2020variational} has extended it to the noisy likelihood case. This approach is an order of magnitude more expensive than the WSABI approach because it essentially involves BQ inside of an optimisation loop for variational inference. This paper employs \gls{wsabil} and its \gls{af} for simplicity and faster computation. Still, our approach is a natural extension of BQ methods and compatible with the above advances. (e.g. changing the RCHQ kernel into the prospective uncertainty sampling AF for VMBC)

\paragraph{Kernel quadrature}
There are a number of \gls{kq} algorithms from
herding/optimisation \citep{chen2010super, bach2012on, huszar2012optimally} to random sampling
\citep{bach2017on, belhadji2019kernel}.
In the context of \gls{bq}, \gls{fwbq} \citep{briol2015frank} using kernel herding has been proposed. This method proposes to do \gls{bq} with the points given by herding,
but the guaranteed exponential convergence $\mathcal{O}(\exp(-cn))$ is limited to finite-dimensional \gls{rkhs},
which is not the case in our setting.
For general kernels, the convergence rate drops to $\mathcal{O}(1/\sqrt{n})$ \citep{bach2012on}.
Recently, a random sampling method with a good convergence rate was proposed for infinite-dimensional \gls{rkhs}
\citep{hayakawa2021positively}.
Based on Carathéodory's theorem for the convex hull,
it efficiently calculates a reduced measure for a larger empirical measure.
In this paper, we call it \gls{rchq} and use it in combination with the \gls{bq} methods. (See Supplementary)

\section{Proposed Method: BASQ}
\subsection{General idea}

We now introduce our algorithm, named \gls{basq}.

\paragraph{Kernel recombination for batch selection}
Batch \gls{wsabi} \citep{wagstaff2018batch} selects batch samples based on the \gls{af}, taking samples having the maximum AF values greedily via gradient-based optimisation with multiple starting points. The computational cost of this sampling scheme increases exponentially with the batch size and/or the problem dimension. Moreover, this method is often only locally optimal \citep{wilson2018maximizing}. We adopt a scalable, gradient-free optimisation via \gls{kq} algorithm based on kernel recombination  \citep{hayakawa2021positively}. Surprisingly, Husz\'{a}r et al. \citep{huszar2012optimally} pointed out the equivalence between \gls{bq} and \gls{kq} with optimised weights. \gls{kq} can select $n$ samples from the $N$ candidate samples $\textbf{X}_\text{rec}$ to efficiently reduce the worst-case error. The problem in batch \gls{bq} is selecting $n$ samples from the probability measure $\pi(x)$ that minimises integral variance. When subsampling $N$ candidate samples $\textbf{X}_\text{rec} \sim \pi(x)$, we can regard this samples $\textbf{X}_\text{rec}$ as an empirical measure $\pi_\text{emp}(x)$ approximating the true probability measure $\pi(x)$ if $n \ll N$. Therefore, applying \gls{kq} to select $n$ samples that can minimise $\mathbb{V}\text{ar}[ \ell(x) \pi_\text{emp}(x)]$ is equivalent to selecting $n$ batch samples for batch \gls{bq}. As more observations make surrogate model $\ell(x)$ more accurate, the empirical integrand model $\ell(x)\pi_\text{emp}(x)$ approaches to the true integrand model $\ell_\text{true}(x) \pi(x)$. This subsampling scheme allows us to apply any \gls{kq} methods for batch \gls{bq}. However, such a dual quadrature scheme tends to be computationally demanding. Hayakawa et al. \citep{hayakawa2021positively} proposed an efficient algorithm based on kernel recombination, \gls{rchq}, which automatically returns a sparse set of $n$ samples based on Carathéodory's theorem. The computational cost of batch size $n$ for our algorithm, \gls{basq}, is lower than
$\mathcal{O}(NM + M^2\log n + Mn^2\log(N/n))$ \citep{hayakawa2021positively}.

\paragraph{Alternately subsampling}
The performance of \gls{rchq} relies upon the quality of a predefined kernel. Thus, we add \gls{bq} elements to \gls{kq} in return; making \gls{rchq} an online algorithm. Throughout the sequential batch update, we pass the \gls{bq}-updated kernels to \gls{rchq}.\footnote{
For updating kernel, the kernel to be updated is $C_\textbf{y}$, not the kernel $K$. The kernel $K$ just corresponds to the {\it prior} belief
in the distribution of $\ell$, so once we have observed the samples $\textbf{X}$ (and $\textbf{y}$), the variance to be minimised becomes $C_\textbf{y}$.
} This enables \gls{rchq} to exploit the function shape information to select the best batch samples for minimising $\mathbb{V}\text{ar}[ \ell(x) \pi_\text{emp}(x)]$. This corresponds to the batch maximisation of the model evidence for $\ell(x)$. Then, \gls{bq} optimises the hyperparameters based on samples from true likelihood  $\ell_\text{true}(x)$, which corresponds to an optimised kernel preset for \gls{rchq} in the next round. These alternate updates characterise our algorithm, \gls{basq}. (See Supplementary)

\paragraph{Importance sampling for uncertainty}
We added one more \gls{bq} element to \gls{rchq}; \textit{uncertainty sampling}. \gls{rchq} relies upon the quality of subsamples from the prior distribution. However, sharp, multimodal, or high-dimensional likelihoods make it challenging to find prior subsamples that overlap over the likelihood distribution. This typically happens in Bayesian inference when the likelihood is expressed as the product of Gaussians, which gives rise to a very sharp likelihood peak. The likelihood of big data tends to become multimodal \citep{roberts2013gaussian}. Therefore, we adopt \textit{importance sampling}, gravitating subsamples toward the meaningful region, and correct the bias via its weights. We propose a mixture of prior and an \gls{af} as a guiding \textit{proposal distribution}. The prior encourages global (rather than just local) optimisation, and the \gls{af} encourages sampling from uncertain areas for faster convergence. However, sampling from \gls{af} is expensive. We derive an efficient sparse Gaussian mixture sampler. Moreover, introducing square-root warping \citep{gunter2014sampling} enables the sampling distribution to factorise, yielding faster sampling.

\paragraph{Summary of contribution}
\label{sec:summary}
\begin{figure}
  %\centering
  \includegraphics[width=1.0\textwidth,center]{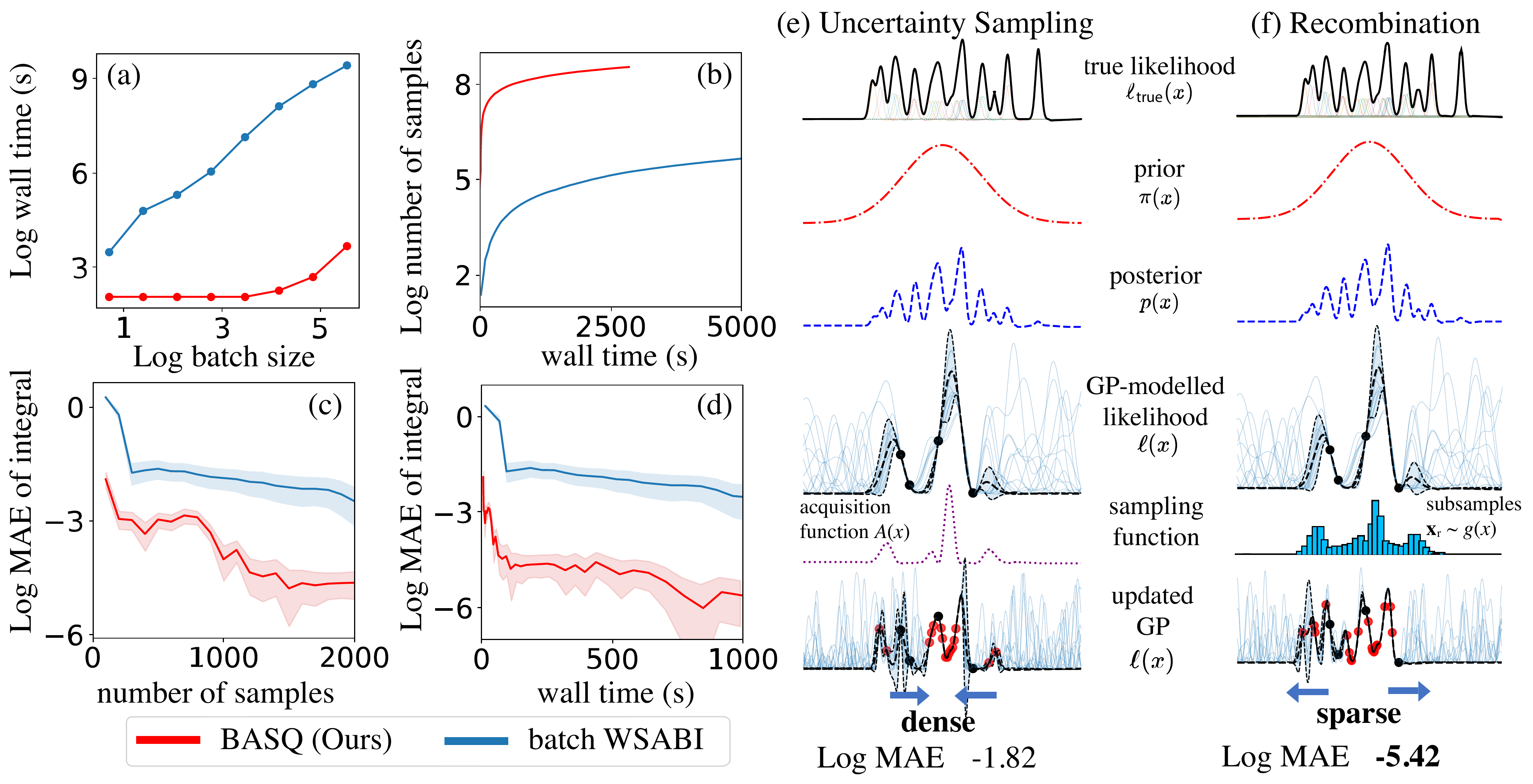}%
  \caption{Performance comparison of our algorithm \gls{basq} against batch \gls{wsabi} \citep{wagstaff2018batch}. All evaluation was performed with the likelihood of a mixture of N-dimensional Gaussians. (a), (b), (c), (d) 10-dimensional Gaussians, (e), (f) univariate Gaussians.}
  \label{fig:BASQvsWSABI}
\end{figure}
We summarised the key differences between batch \gls{wsabi} and \gls{basq} in Figure ~\ref{fig:BASQvsWSABI}.\footnote{We randomly varied the number of components between 10 and 15, setting their variance uniformly at random between 1 and 4, and setting their means uniformly at random within the box bounded by [-3,3] in all dimensions. The weights of Gaussians were randomly generated from a uniform distribution, but set to be one after integration. \gls{mae} was adopted for the evaluation of integral estimation.} (a) shows that \gls{basq} is more scalable in batch size. (b) clarifies that \gls{basq} can sample 10 to 100 times as many samples in the same time budget as \gls{wsabi}, supported by the efficient sampler and \gls{rchq}. (c) states the convergence rate of \gls{basq} is faster than \gls{wsabi}, regardless of computation time. (d) demonstrates the combined acceleration in wall time. While the batch \gls{wsabi} reached $10^{-1}$ after 1,000 seconds passed, \gls{basq} was within seconds. (e) and (f), visualised how \gls{rchq} selects sparser samples than batch \gls{wsabi}. This clearly explains that gradient-free kernel recombination is better in finding the global optimal than multi-start optimisation. These results demonstrate that we were able to combine the merits of \gls{bq} and \gls{kq} (see Supplementary). We further tested with various synthetic datasets and real-world tasks in the fields of lithium-ion batteries and material science. Moreover, we mathematically analyse the convergence rate with proof in Section \ref{sec:convergence}.

%A Python implementation of our method is available online at \href{https://github.com/ma921/BASQ/}{https://github.com/ma921/BASQ/}

\subsection{Algorithm}
\newcommand{\codecomment}[1]{\hspace{\fill}\rlap{\# #1}\phantom{Define $n-1$ test functions via Nyström}}

\begin{table}
  \caption{BASQ algorithm}
  \label{algorithm}
  \centering
  \begin{adjustwidth}{-1in}{-1in}
  \begin{center}
  \begin{tabular}{l}
    \toprule
    \textbf{Algorithm 1}: Bayesian Alternately Subsampled Quadrature (BASQ)
    \\%
    \midrule%
    ~~~~~~\textbf{Notation}: $x_{\text{init}}$: initial guess, $k$: a convergence criterion, \\
    \hspace{55pt} $n$: batch size, $M$: Nyström sample size, $N$: recombination sample size, \\
    \hspace{55pt} $\ell$: GP-modelled likelihood, $\ell_{\text{true}}$: true likelihood, \\
    \hspace{55pt} $\pi(x)$: prior, $p(x)$: posterior, $A(x)$: \gls{af}, $K$: kernel, \\
    \hspace{55pt} $S_\text{nys}$, $S_\text{rec}$: samplers for Nyström and recombination, respectively \vspace{2mm}\\
    ~~~~~~\textbf{Input}: prior $\pi(x)$, true likelihood function $\ell_{\text{true}}$\\%
    ~~~~~~\textbf{Output}: posterior $p(x)$, the mean and variance of model evidence $\mathbb{E}[Z \,|\,\textbf{y}], \text{var}[Z \,|\, \textbf{y}]$ \vspace{3mm}\\%
    ~~1: $C^L_{\textbf{y}}, A(x) \leftarrow \text{InitialiseGPs}(x_{\text{init}})$ \codecomment{Initialise GPs with initial guess}\\%
    ~~2: $S_\text{nys}, S_\text{rec} \leftarrow \text{SetSampler}(A(x), \pi(x))$ \codecomment{Set samplers}\\%
    ~~3: \textbf{while} $\text{var}[Z \,|\, \textbf{y}] > k$:\\%
    ~~4: ~~~~~$\textbf{X}_\text{nys} \sim S_\text{nys}(M)$ \codecomment{Samples $M$ points for test function $\varphi$}\\%
    ~~5: ~~~~~$(\textbf{w}_\text{rec}, \textbf{X}_\text{rec})
    \sim S_\text{rec}(N)$ \codecomment{Samples $N$ points for recombination}\\%
    ~~6: ~~~~~$\varphi_1,...,\varphi_{n-1}
        \leftarrow \text{Nyström}(\textbf{X}_\text{nys}, C^L_{\textbf{y}})$ \codecomment{Define $n-1$ test functions}\\%
    ~~7: ~~~~~\textbf{solve a kernel recombination problem}\\
    ~~8: ~~~~~~~~~~Find an $n$-point subset
    $\textbf{X}_\text{quad}\subset\textbf{X}_\text{rec}$
    and $\textbf{w}_\text{quad}\ge \bm{0}$
    \\%
    ~~9: ~~~~~~~~~~s.t. $\textbf{w}_\text{quad}^\top \varphi_i(\textbf{X}_\text{quad})=
        \textbf{w}_\text{rec}^\top \varphi_i(\textbf{X}_\text{rec})$,
        $\textbf{w}_\text{quad}^\top\bm1 = \textbf{w}_\text{rec}^\top\bm1$\\%
    10: ~~~~~~~~~~\textbf{return} $\textbf{X}_\text{quad}$
    \codecomment{The sparse set of $n$ samples}\\%
    11: ~~~~~$\textbf{y} = \text{Parallel}(\ell_{\text{true}}(\textbf{X}_\text{quad}))$ \codecomment{Parallel computing of likelihood}\\%
    12: ~~~~~$K \leftarrow \text{Update}(\textbf{X}_\text{quad}, \textbf{y})$ \codecomment{Train GPs}\\%
    13: ~~~~~$C^L_{\textbf{y}}, A(x) \leftarrow \text{OptHypersThenUpdate}(K)$
    \codecomment{Type II MLE optimisation}\\
    14: ~~~~~$S_\text{nys}, S_\text{rec} \leftarrow \text{ResetSampler}(A(x), \pi(x))$ \codecomment{Reset samplers with the updated $A(x)$}\\%
    15: ~~~~~$\mathbb{E}[Z \,|\,\textbf{y}], \text{var}[Z \,|\, \textbf{y}] \leftarrow \text{BayesQuad}(m^L_{\textbf{y}}, C^L_{\textbf{y}}, \pi(x))$ \codecomment{Calculate via Eqs. \eqref{eq:BQ} and \eqref{eq:WSABI}}\\
    16: \textbf{return} $p(x), \mathbb{E}[Z \,|\,\textbf{y}], \text{var}[Z \,|\, \textbf{y}]$\\

    \bottomrule
  \end{tabular}
  \end{center}
  \end{adjustwidth}
\end{table}

Table ~\ref{algorithm} illustrates the pseudo-code for our algorithm. Rows 4 - 10 correspond to \gls{rchq}, and rows 11 - 15 correspond to \gls{bq}. We can use the variance of the integral $\mathbb{V}\text{ar}[Z | \textbf{y}]$ as a convergence criterion. For hyperparameter optimisation, we adopt the type-II \gls{mle} to optimise hyperparameters via L-BFGS \citep{byrd1995limited} for speed.

\paragraph{Importance sampling for uncertainty}
Lemma 1 in the supplementary proves the optimal upper bound of the proposal distribution $g(x) \approx \sqrt{K(x, x)}f(x) = \sqrt{C^L_\textbf{y}(x, x)}f(x)$, where $f(x) := \pi(x)$. However, sampling from square-root variance is intractable, so we linearised to $g(x) \approx 0.5(1 + C^L_\textbf{y}(x, x))f(x)$. To correct the linearisation error, the coefficient 0.5 was changed into the hyperparameter $r$, which is defined as follows:
\begin{align}
 g(x) &= (1 - r) f(x) + r \tilde A(x), \quad 0 \leq r \leq 1 \label{eq:IS-a}\\
 \tilde A(x) &= \frac{C^L_{\textbf{y}}(x, x)\pi(x)}{\int C^L_{\textbf{y}}(x, x)\pi(x) dx}, \label{eq:IS-b}\\
 \text{w}_\text{IS}(x) &= f(x) / g(x), \label{eq:IS-c}
 \end{align}\label{eq:IS}
where $\text{w}_\text{IS}$ is the weight of the importance sampling. While $r=1$ becomes the pure uncertainty sampling, $r=0$ is the vanilla \gls{mc} sampler.

\paragraph{Efficient sampler}
Sampling from $A(x)$, a mixture of Gaussians, is expensive, as some mixture weights are negative, preventing the usual practice of weighted sampling from each Gaussian. As the warped kernel $C^L_{\textbf{y}}(x,x)$ is also computationally expensive,
we adopt a \textit{factorisation trick}:
\begin{align}
    Z = \int \ell(x) \pi(x)\dd x = \alpha + \frac12 \int |\tilde{\ell}(x)|^2 \pi(x)\dd x \approx \alpha + \frac12 \int |\tilde{\ell}(x)| f(x) \dd x,
\end{align}
where we have changed the distribution of interest to $f(x)=|\tilde{m_{\textbf{y}}}(x)|\pi(x)$. This is doubly beneficial. Firstly, the distribution of interest $f(x)$ will be updated over iterations. The previous $f(x) = \pi(x)$ means the subsample distribution eventually obeys prior, which is disadvantageous if the prior does not overlap the likelihood sufficiently. On the contrary, the new $f(x)$ narrows its region via $|\tilde{m_{\textbf{y}}}(x)|$. Secondly, the likelihood function changed to $|\tilde{\ell}(x)|$, thus the kernels shared with \gls{rchq} changed into cheap warped kernel $\tilde C_{\textbf{y}}(x,x)$. This reduces the computational cost of \gls{rchq}, and the sampling cost of $A(x)$. Now $A(x)=\tilde{C_{\textbf{y}}}(x)\pi(x)$, which is also a Gaussian mixture, but the number of components is significantly lower than the original \gls{af} \eqref{eq:IS-b}. As $\tilde{C_{\textbf{y}}}(x)$ is positive, the positive weights of the Gaussian mixture should cover the negative components. Interestingly, in many cases, the positive weights vary exponentially, which means that limited number of components dominate the functional shape. Thus, we can ignore the trivial components for sampling.\footnote{Negative elements in the matrices only exist in $K(\textbf{X}, \textbf{X})^{-1}$, which can be drawn from the memory of the GP regression model without additional calculation. The number of positive components is half of the matrix on average, resulting in $\mathcal{O}(n^2/2)$. Then, taking the threshold via the inverse of the recombination sample size $N$, the number of components becomes $n_\text{comp} \ll n^2$, resulting in sampling complexity $\mathcal{O}(n^2/2 + n_\text{comp}N)$.} Then we adopt \gls{smc} \citep{kitagawa1993monte} to sample $A(x)$. We have a highly-calibrated proposal distribution of sparse Gaussian mixture, leading to efficient resampling from real $A(x)$ (see Supplementary).

\paragraph{Variants of proposal distribution}
Although \eqref{eq:IS-a} has mathematical motivation, sometimes we wish to incorporate prior information not included in the above procedure. We propose two additional ``biased'' proposal distributions. The first case is where we know both the maximum likelihood points and the likelihood's unimodality. This is typical in Bayesian inference because we can obtain (sub-)maximum points via a \gls{map} estimate. In this case, we know exploring around the perfect initial guess is optimal rather than unnecessarily exploring an uncertain region. Thus, we introduce the \gls{igb} proposal distribution, $g_{\text{IGB}}(x)$. This is written as $g_{\text{IGB}}(x) = (1-r)\pi(x) + r \sum_{i=1} w_{i, \text{IGB}}\mathcal{N}(x; X_i, \textbf{W})$, where $w_{i, \text{IGB}} = \left\{ 0 \ \ \text{if} \ \ y_i \leq 0, \ \ \text{else} \ \ 1  \right\}$, $X_i \in \textbf{X}$. This means exploring only the vicinity of the observed data $\textbf{X}$. The second case is where we know the likelihood is multimodal. In this case, determining all peak positions is most beneficial. Thus more explorative distribution is preferred. As such, we introduce the \gls{ub} proposal distribution, $g_{\text{UB}}(x)$. This is written as $g_{\text{UB}}(x) = A(x)$, meaning pure uncertainty sampling. To contrast the above two, we term the proposal distribution in Eq. \eqref{eq:IS-a} as \gls{ivr} $g_{\text{IVR}}(x)$.

\section{Experiments}
\label{sec:experiments}
Given our new model \gls{basq}, with three variants of the proposal distribution, \gls{ivr}, \gls{igb}, and \gls{ub}, we now test for speed against \gls{mc} samplers and batch \gls{wsabi}. We compared with three \gls{ns} methods \citep{skilling2006nested, feroz2009multinest, higson2019dynamic, buchner2016statistical, buchner2019collaborative}, coded with \citep{speagle2020dynesty, buchner2021ultranest}. According to the review \citep{buchner2021nested}, MLFriends  is the state-of-the-art \gls{ns} sampler to date. The code is implemented based on \citep{wagstaff2019bayesian, gpy2012gpy, harris2020array, virtanen2020scipy, gurobi2022gurobi, gardner2018gpytorch, berahas2016multi, balandat2020botorch},
and code around kernel recombination \citep{cosentino2020randomized,hayakawa2021positively}
with additional modification.
All experiments on synthetic datasets were averaged over 10 repeats, computed in parallel with multicore CPUs, without GPU for fair comparison.\footnote{Performed on MacBook Pro 2019, 2.4 GHz 8-Core Intel Core i9, 64 GB 2667 MHz DDR4} The posterior distribution of \gls{ns} was estimated via \gls{kde} with weighted samples \citep{gsbert2003weighted}. For maximum speed performance, batch size was optimised for each method in each dataset, in fairness to the competitors. Batch \gls{wsabi} needs to optimise batch size to balance the likelihood query cost and sampling cost, because sampling cost increases rapidly with batch size, as shown in Figure ~\ref{fig:BASQvsWSABI}(a). Therefore, it has an optimal batch size for faster convergence. By wall time cost, we exclude the cost of integrand evaluation; that is, the wall time cost is the overhead cost of batch evaluation. Details can be found in the Supplementary.

\subsection{Synthetic problems}
We evaluate all methods on three synthetic problems. The goal is to estimate the integral and posterior of the likelihood modelled with the highly multimodal functions. Prior was set to a two-dimensional multivariate normal distribution, with a zero mean vector, and covariance whose diagonal elements are 2. The optimised batch sizes for each methods are \gls{basq}: 100, batch \gls{wsabi}: 16. The synthetic likelihood functions are cheap (0.5 ms on average). This is advantageous setting for \gls{ns}: Within 10 seconds, the batch \gls{wsabi}, \gls{basq}, and \gls{ns} collected 32, 600, 23225 samples, respectively. As for the metrics, posterior estimation was tested with \gls{kl} upon random 10,000 samples from true posterior. Evidence was evaluated with \gls{mae}, and ground truth was derived analytically.

\paragraph{Likelihood functions}
\textit{Branin-Hoo} \citep{jones2001taxonomy} is 8 modal function in two-dimensional space. \textit{Ackley}  \citep{surjanovic2022virtual} is a highly multimodal function with point symmetric periodical peaks in two-dimensional space. \textit{Oscillatory function} \citep{Genz1984testing} is a highly multimodal function with reflection symmetric periodical peaks of highly-correlated ellipsoids in two-dimensional space.

\subsection{Real-world dataset}
We consider three real-world applications with expensive likelihoods, which are simulator-based and hierarchical GP. We adopted the empirical metric due to no ground truth. For the posterior, we can calculate the true conditional posterior distribution along the line passing through ground truth parameter points. Then, evaluate the posterior with \gls{rmse} against 50 test samples for each dimension. For integral, we compare the model evidence itself. Expensive likelihoods makes the sample size per wall time amongst the methods no significant difference, whereas rejection sampling based \gls{ns} dismiss more than 50\% of queried samples. The batch sizes are \gls{basq}: 32, batch \gls{wsabi}: 8. (see Supplementary)

\paragraph{Parameter estimation of the lithium-ion battery simulator}: The simulator is the SPMe \citep{marquis2019asymptotic}, \footnote{SPMe code used was translated into Python from MATLAB \citep{bizeray2016spectral, bizeray2015lithium}. This open-source code is published under the BSD 3-clause Licence. See more information on \citep{bizeray2016spectral}} estimating 3 simulation parameters at a given time-series voltage-current signal (the diffusivity of lithium-ion on the anode and cathode, and the experimental noise variance). Prior is modified to log multivariate normal distribution from \citep{aitio2020bayesian}. Each query takes 1.2 seconds on average. 

\paragraph{Parameter estimation of the phase-field model}: The simulator is the phase-field model \citep{marquis2019asymptotic}, \footnote{Code used was from \citep{koyama2019computational, aitio2021predicting}. All rights of the code are reserved by the authors. Thus, we do not redistribute the original code.} estimating 4 simulation parameters at given time-series two-dimensional morphological image (temperature, interaction parameter, Bohr magneton coefficient, and gradient energy coefficient). Prior is a log multivariate normal distribution. Each query takes 7.4 seconds on average.

\paragraph{Hyperparameter marginalisation of hierarchical GP model} The hierarchical \gls{gp} model was designed for analysing the large-scale battery time-series dataset from solar off-grid system field data \citep{aitio2021predicting}.$^8$ For fast estimation of parameters in each GP, the recursive technique \citep{sarkka2012infinite} is adopted. The task is to marginalise 5 \gls{gp} hyperparameters at given hyperprior, which is modified to log multivariate normal distribution from \citep{aitio2021predicting}. Each query takes 1.1 seconds on average.

\subsection{Results}
\begin{figure}
  %\centering
  \includegraphics[width=1.00\textwidth,center]{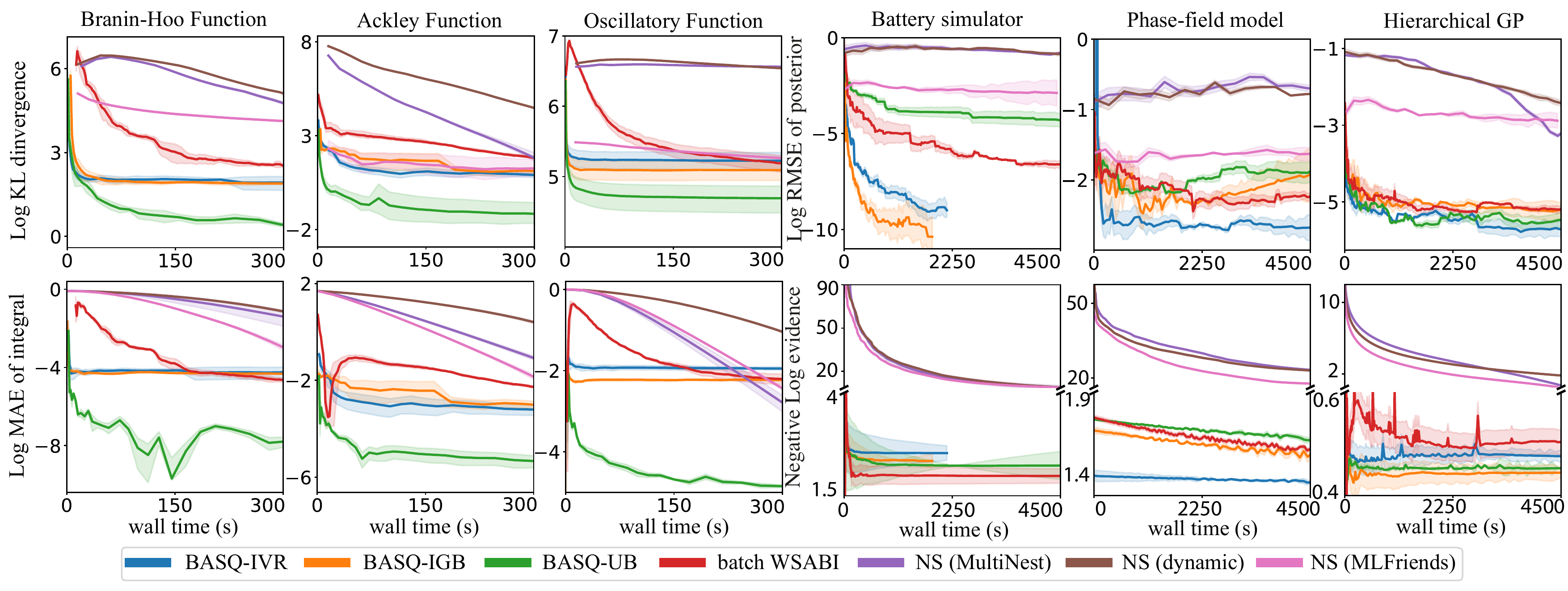}%
  \caption{Time in seconds vs. \gls{kl} divergence for posterior, and \gls{mae} for evidence in the synthetic datasets. Time in seconds vs. \gls{rmse} for posterior and evidence itself in the real-world dataset.}
  \label{fig:benchmarks}
\end{figure}

We find BASQ consistently delivers strong empirical performance, as shown in Figure ~\ref{fig:benchmarks}. On all benchmark problems, BASQ-IVR, IGB, or UB outperform baseline methods except in the battery simulator evidence estimation. The very low-dimensional and sharp unimodal nature of this likelihood could be advantageous for biased greedy batch WSABI, as IGB superiority supports this viewpoint. This suggests that BASQ could be a generally fast Bayesian solver as far as we investigated. In the multimodal setting of the synthetic dataset, BASQ-UB outperforms, whereas IVR does in a simulator-based likelihoods. When comparing each proposal distribution, BASQ-IVR was the performant. Our results support the general use of IVR, or UB if the likelihood is known to be highly multimodal.

\section{Convergence analysis}\label{sec:convergence}
\label{sec:theory}
We analysed the convergence over single iteration on a simplified version of BASQ, which assumes the BQ is modelled with vanilla BQ, without batch and hyperparameter updates.
Note that the kernel $K$ on $\R^d$
in this section
refers to the given covariance kernel of \gls{gp} at each step.
We discuss the convergence of \gls{basq} in one iteration.
We consider the importance sampling:
Let $f$ be a probability density on $\R^d$
and $g$ be another density such that $f = \lambda g$
with a nonnegative function $\lambda$.
Let us call such a pair, $(f, g)$,
a density pair with weight $\lambda$.

We approximate the kernel $K$ with
$K_0 = \sum_{i=1}^{n-1}c_i\varphi_i(x)\varphi_i(y)$.
In general, we can apply the kernel recombination algorithm
\citep{lit12,tch15}
with the weighted sample $(\textbf{w}_\text{rec}, \textbf{X}_\text{rec})$
to obtain a weighted point set $(\textbf{w}_\text{quad}, \textbf{X}_\text{quad})$
of size $n$ satisfying
$
    \textbf{w}_\text{quad}^\top \varphi_i(\textbf{X}_\text{quad})
    = \textbf{w}_\text{rec}^\top \varphi_i(\textbf{X}_\text{rec})
$ ($i=1,\ldots,n-1$) and $
    \textbf{w}_\text{quad}^\top \bm1
    = \textbf{w}_\text{rec}^\top \bm1
$.
By modifying the kernel recombination algorithm,
we can require $\textbf{w}_\text{quad}^\top k_1^{1/2}(\textbf{X}_\text{quad})
    \le \textbf{w}_\text{rec}^\top k_1^{1/2}(\textbf{X}_\text{rec})$,
where $k_1^{1/2}(x) := \sqrt{K(x, x) - K_0(x, x)}$ \citep{hayakawa2021positively}.
We call such $(\textbf{w}_\text{quad}, \textbf{X}_\text{quad})$
a {\it proper} kernel recombination of $(\textbf{w}_\text{rec}, \textbf{X}_\text{rec})$
with $K_0$.\footnote{Note that the inequality constraint on the diagonal value here is only needed for theoretical guarantee,
and skipping it does not reduce the empirical performance \citep{hayakawa2021positively}.}
We have the following guarantee (proved in Supplementary):

\begin{thm}
    Suppose $\int \sqrt{K(x, x)}f(x)\dd x<\infty$,
    \rm$\ell\sim \mathcal{GP}(m, K)$\it,
    and we are given an $(n-1)$-dimensional kernel
    \rm$K_0$\it\
    such that
    \rm$K_1:=K - K_0$\it\
    is also a kernel.
    Let $(f, g)$ be a density pair with weight $\lambda$.
    Let \rm$\textbf{X}_\text{rec}$\it\ be
    an $N$-point independent sample from $g$
    and \rm$\textbf{w}_\text{rec}:=\lambda(\textbf{X}_\text{rec})$\it.
    Then, if \rm$(\textbf{w}_\text{quad}, \textbf{X}_\text{quad})$\it\
    is a proper kernel recombination of \rm$(\textbf{w}_\text{rec}, \textbf{X}_\text{rec})$\it\
    for \rm$K_0$\it, it satisfies
    \rm\begin{equation}
        \mathbb{E}_{\textbf{x}_\text{rec}}\!\left[
        \sqrt{\text{var}[
            Z_f
            \,|\, \textbf{x}_\text{quad}]}\right]
        \le 2\left(\int K_1(x, x)f(x)\dd x\right)^{1/2}
         + \sqrt{\frac{C_{K, f, g}}N},
         \label{eq:guarantee-main}
    \end{equation}\it
    where \rm$Z_f:= \int \ell(x)f(x)\dd x$\it\
    and \rm$C_{K, f, g}
            :=\int K(x, x)\lambda(x)f(x)\dd x
            - \iint K(x, y)f(x)f(y) \dd x\dd y$\it.
\end{thm}

The above approximation has one source of randomness which stems from sampling $N$ points $\textbf{x}_\text{rec}$ from $g$.
One can also apply this estimate
with a random kernel and thereby introduce another source of randomness. In particular, when we use the Nystr{\"o}m approximation for $K_0$
(that ensures $K_1$ is a kernel \citep{hayakawa2021positively}),
then one can show that  $\int K_1(x,x)f(x) \dd x$
can be bounded by
\begin{equation}
    \int K_1(x, x)f(x)\dd x
    \le n\sigma_n + \sum_{m={n+1}}^\infty\sigma_m
    + \mathcal{O}_p\left(\frac{nK_{\max}}{\sqrt{M}}\right),
    \label{eq:nys-ordp}
\end{equation}
where $\sigma_n$ is the $n$-th eigenvalue
of the integral operator $L^2(f)\ni h\mapsto \int
K(\cdot, y)h(y)f(y)\dd x$, $K_{\max} :=\sup_x K(x, x)$.
However, note that unlike Eq. \eqref{eq:guarantee-main},
this inequality only applies with high probability due
to the randomness of $K_0$; see Supplementary for details.

If, for example,
$K$ is a Gaussian kernel on $\R^d$
and $f$ is a Gaussian distribution,
we have $\sigma_n = \mathcal{O}(\exp(-cn^{1/d}))$
for some constant $c>0$
(see Supplementary).
So in \eqref{eq:guarantee-main}
we also achieve an empirically exponential rate when $N \gg C_{K, f, g}$ .
RCHQ works well with a moderate $M$ in practise.
Note that unlike the previous analysis \citep{kan19},
we do not have to assume that the space is compact.
\footnote{
The rate $N^{-1/2}$ for the expected integral variance on non-compact sets can also be found in \cite{graf2013efficient} [Corollary 2.8]. In the case of Gaussian integration distribution and the Gaussian kernel exponential rates of convergence of the BQ integral variance can be found \cite{belhadji2019kernel} [Theorem 1], \cite{kuo2012gauss} [Theorem 4.1], \cite{Kuo2017multivariate} [Theorem 3.2], \cite{karvonen2021integration} [Theorems 2.5 and 2.10]
}

\section{Discussion}
\label{sec:future}
We introduced a batch \gls{bq} approach, \gls{basq}, capable of simultaneous calculation of both model evidence and posteriors. \gls{basq} demonstrated faster convergence (in wall-clock time) on both synthetic and real-world datasets, when compared against existing \gls{bq} approaches and state-of-the-art \gls{ns}. Further, mathematical analysis shows the possibility to converge exponentially-fast under natural assumptions. As the BASQ framework is general-purpose, this can be applied to other active learning GP-based applications, such as Bayesian optimisation \citep{kathuria2016batched}, dynamic optimisation like control \citep{deisenroth2009gaussian}, and probabilistic numerics like ODE solvers \citep{hennig2015probabilistic}. Although it scales to the number of data seen in large-scale GP experiments, practical BASQ usage is limited to fewer than 16 dimensions (similar to many \gls{gp}-based algorithms). However, RCHQ is agnostic to the input space, allowing quadrature in manifold space. An appropriate latent variable warped GP modelling, such as GPLVM \citep{lawrence2003gaussian}, could pave the way to high dimensional quadrature in future work. In addition, while WSABI modelling limits the kernel to a squared exponential kernel, RCHQ allows to adopt other kernels or priors without a bespoke modelling BQ models. (See Supplementary).
As for the mathematical proof, we do not incorporate batch and hyperparameter updates, which should be addressed in future work. The generality of our theoretical guarantee with respect to kernel and distribution should be useful for extending the analysis to the whole algorithm.

\begin{ack}
We thank Saad Hamid and Xingchen Wan for the insightful discussion of Bayesian quadrature, and Antti Aitio and David Howey for fruitful discussion of Bayesian inference for battery analytics and for sharing his codes on the single particle model with electrolyte dynamics, and hierarchical GPs. We would like to thank Binxin Ru, Michael Cohen, Samuel Daulton, Ondrej Bajgar, and anonymous reviewers for their helpful comments about improving the paper. Masaki Adachi was supported by the Clarendon Fund, the Oxford Kobe Scholarship, the Watanabe Foundation, the British Council Japan Association, and Toyota Motor Corporation.
Harald Oberhauser was supported by the DataSig Program [EP/S026347/1], the Hong Kong Innovation and Technology Commission (InnoHK Project CIMDA), and the Oxford-Man Institute. Martin Jørgensen was supported by the Carlsberg Foundation.
\end{ack}

\medskip
\small

%\bibliographystyle{unsrtnat}
%\setcitestyle{numbers,open={[},close={]},citesep={,}}
\bibliographystyle{plainurl}
\sloppy
\bibliography{reference}

%%%%%%%%%%%%%%%%%%%%%%%%%%%%%%%%%%%%%%%%%%%%%%%%%%%%%%%%%%%%

\newpage
\appendix

\section{Convergence analysis}\label{sec:app-nys}
\subsection{Proof of Theorem 1}
We provide the proof of the following theorem
given in the main text.
\begin{thm}\label{thm:recall-main}
    Suppose $\int \sqrt{K(x, x)}f(x)\dd x<\infty$,
    \rm$\ell\sim \mathcal{GP}(m, K)$\it, 
    and we are given an $(n-1)$-dimensional kernel
    \rm$K_0$\it\
    such that
    \rm$K_1:=K - K_0$\it\ 
    is also a kernel.
    Let $(f, g)$ be a density pair with weight $\lambda$.
    Let \rm$\textbf{x}_\text{rec}$\it\ be
    an $N$-point independent sample from $g$
    and \rm$\textbf{w}_\text{rec}:=\lambda(\textbf{x}_\text{rec})$\it.
    Then, if \rm$(\textbf{w}_\text{quad}, \textbf{x}_\text{quad})$\it\
    is a proper recombination of \rm$(\textbf{w}_\text{rec}, \textbf{x}_\text{rec})$\it\
    for \rm$K_0$\it, it satisfies
    \rm\begin{equation}
        \mathbb{E}_{\textbf{x}_\text{rec}}\!\left[
        \sqrt{\text{var}[
            Z_f 
            \,|\, \textbf{x}_\text{quad}]}\right]
        \le 2\left(\int K_1(x, x)f(x)\dd x\right)^{1/2}
         + \sqrt{\frac{C_{K, f, g}}N}
         %\label{eq:guarantee-main}
    \end{equation}\it
    where \rm$Z_f:= \int \ell(x)f(x)\dd x$\it\ 
    and \rm$C_{K, f, g}
            :=\int K(x, x)\lambda(x)f(x)\dd x
            - \iint K(x, y)f(x)f(y) \dd x\dd y$\it.
\end{thm}

Recall $\H$ is the RKHS given by
the kernel $ K$.
As the kernel satisfies $\int\sqrt{K(x, x)}f(x)\dd x < \infty$,
the mean embedding
\begin{equation}
    \mu_K(f):=\int f(x) K(x, \cdot)\dd x
    \label{eq:km-embed}
\end{equation}
is a well-defined element of $\H$.
We first discuss its approximation via importance sampling.
\begin{lem}\label{lem:is}
    Let $f$ be a probability density on $\R^d$
    and $g$ be another density such that $f = \lambda g$
    with a nonnegative function $\lambda$.
    Let \rm$\textbf{x}_\text{rec}$\it\ be
    an $N$-point independent sample from $g$
    and \rm$\textbf{w}_\text{rec} = \lambda(\textbf{x}_\text{rec})$\it\ be the weights.
    If we define \rm$\mu_r:= \frac1N\textbf{w}_\text{rec}^\top  K(\textbf{x}_\text{rec}, \cdot)$\it\ then it satisfies
    \rm\begin{subequations}
        \begin{eqnarray*}
            &\mathbb{E}[\lVert \mu_K(f) - \mu_r \rVert_\H^2]
            = \frac1NC_{ K, f, g}\\
            &\text{where}\quad
            C_{ K, f, g}
            =\int  K(x, x)\lambda(x)f(x)\dd x
            - \iint  K(x, y)f(x)f(y) \dd x\dd y
            \label{eq:c_kfg}
        \end{eqnarray*}
    \end{subequations}\it
    Furthermore,
    the choice \rm$g(x) \propto \sqrt{ K(x, x)}f(x)$\it\
    minimises \rm$C_{ K, f, g}$, if
    \rm$\lambda =  K(x, x)^{-1/2}$\it\
    is well-defined.
\end{lem}
\begin{proof}
    Let $\textbf{x}_\text{rec} = (X_1, \ldots, X_N)$,
    so $\mu_r = \frac1N\sum_{i=1}^N\lambda(X_i)
     K(X_i, \cdot)$.
    From \eqref{eq:km-embed},
    we have
    \begin{align}
        \lVert \mu_K(f) - \mu_r \rVert_\H^2
        &= \lVert \mu_K(f) \rVert_\H^2
        - 2 \langle \mu_K(f), \mu_r \rangle_\H
        + \lVert \mu_r \rVert_\H^2\\
        &=\iint  K(x, y)f(x)f(y)\dd x\dd y
        -\frac2N\sum_{i=1}^N \int K(x, X_i)
        f(x)\lambda(X_i)\dd x\\
        &\quad + \frac1{N^2} \sum_{i,j=1}^N  K(X_i, X_j)
        \lambda(X_i)\lambda(X_j).
    \end{align}
    We have
    \begin{equation}
        \mathbb{E}\!\left[\int K(x, X_i)f(x)\lambda(X_i)\dd x
        \right]
        = \iint  K(x, y)f(x)\lambda(y)g(y)\dd x\dd y
        = \iint  K(x, y)f(x)f(y)\dd x\dd y
    \end{equation}
    and for $i\ne j$
    \begin{equation}
        \mathbb{E}\!
        \left[ K(X_i, X_j)\lambda(X_i)\lambda(X_j)
        \right]
        = \iint  K(x, y)\lambda(x)\lambda(y)g(x)g(y)
        \dd x\dd y
        = \iint  K(x, y)f(x)f(y)\dd x\dd y,
    \end{equation}
    so we in total have
    \begin{align}
        \mathbb{E}[\lVert \mu_K(f) - \mu_r \rVert_\H^2]
        &=
        \frac1{N^2}\sum_{i=1}^N
        \mathbb{E}[ K(X_i, X_i)\lambda(X_i)^2]
        - \frac1N\iint  K(x, y)f(x)f(y)\dd x\dd y \\
        &= \frac1N\left(\int  K(x, x)\lambda(x)f(x)\dd x
        - \iint  K(x, y)f(x)f(y)\dd x\dd y\right)
        = \frac{C_{ K,f,g}}N.
    \end{align}
    We next show the optimality of
    $g(x)\approx \sqrt{ K(x, x)}f(x)$.
    It suffices to consider when $\int  K(x, x)
    \lambda(x)f(x)\dd x$ is minimised
    as the second term is independent of $g$.
    From the Cauchy-Schwarz, we have
    \begin{equation}
        \int  K(x, x)\lambda(x)f(x)\dd x
        = \int  K(x, x)\lambda(x)f(x)\dd x
        \int \frac{f(x)}{\lambda(x)}\dd x
        \ge
        \left(\int \sqrt{ K(x, x)}f(x)\dd x\right)^2,
    \end{equation}
    and the equality is satisfied if
    $g(x)=\frac{f(x)}{\lambda(x)}\propto
    \sqrt{ K(x, x)}f(x)$.
\end{proof}

\begin{proof}[Proof of Theorem \ref{thm:recall-main}]
Let $(\textbf{w}_\text{quad}, \textbf{x}_\text{quad})$ be a proper recombination of $(\textbf{w}_\text{rec}, \textbf{x}_\text{rec})$,
and let $Q_n$ be the quadrature formula given by points $\textbf{x}_\text{quad}$ and weights $\frac1N\textbf{w}_\text{quad}$,
i.e,
$Q(h):=\frac1N\textbf{w}_\text{quad}^\top h(\textbf{x}_\text{quad})$.
We also define $\mu_n:=\frac1N\textbf{w}_\text{quad}^\top
 K(\textbf{x}_\text{quad},\cdot)$.

A well-known fact is that the worst-case error of $Q_n$
(with respect to $f$ here)
$\text{wce}(Q_n)
    = \sup_{\lVert h\rVert \le 1}
    \lvert Q_n(h) - \int h(x)f(x)\dd x\rvert$
satisfies
$\text{wce}(Q_n)
= \lVert \mu_K(f) - \mu_n\rVert_\H$
for a kernel satisfying
$\int \sqrt{K(x, x)}f(x)\dd x < \infty$
\citep{gretton2006kernel,sriperumbudur2010hilbert}.
By using this and
the relation between Bayesian quadrature and kernel quadrature in the main text,
we have
\begin{align}
    \sqrt{\text{var}[Z_f\,|\,\textbf{x}_\text{quad}]}
    \le \text{wce}(Q_n)
    &\le \lVert \mu_K(f) - \mu_r\rVert_\H
    + \lVert \mu_r - \mu_n\rVert_\H
\end{align}
From Lemma \ref{lem:is} we have
$\mathbb{E}[\lVert \mu_K(f) - \mu_r\rVert_\H]\le \mathbb{E}
[\lVert \mu_K(f) - \mu_r\rVert_\H^2]^{1/2}
= \sqrt{C_{ K, f, g}/N}$,
so it now suffices to show
\begin{equation}
    \mathbb{E}_{\textbf{x}_\text{rec}}[\lVert \mu_r - \mu_n\rVert_\H]
    \le 2\left(\int  K_1(x, x) f(x)\dd x\right)^{1/2}.
    \label{eq:wtp-gen}
\end{equation}

We first have
\begin{align}
    \lVert \mu_r - \mu_n\rVert_\H^2
    &= \frac1{N^2}\left(
    \textbf{w}_\text{rec}^\top  K(\textbf{x}_\text{rec}, \textbf{x}_\text{rec})\textbf{w}_\text{rec} -
    2\,\textbf{w}_\text{rec}^\top  K(\textbf{x}_\text{rec}, \textbf{x}_\text{quad}) \textbf{w}_\text{quad}
    + \textbf{w}_\text{quad}^\top  K(\textbf{x}_\text{quad}, \textbf{x}_\text{quad})\textbf{w}_\text{quad}
    \right),
\end{align}
and from the recombination property we also have
\begin{equation}
    \textbf{w}_\text{rec}^\top  K_0(\textbf{x}_\text{rec}, \textbf{x}_\text{rec})\textbf{w}_\text{rec} -
    2\,\textbf{w}_\text{rec}^\top  K_0(\textbf{x}_\text{rec}, \textbf{x}_\text{quad}) \textbf{w}_\text{quad}
    + \textbf{w}_\text{quad}^\top  K_0(\textbf{x}_\text{quad}, \textbf{x}_\text{quad})\textbf{w}_\text{quad} = 0,
\end{equation}
which follows from
the fact that $(\textbf{w}_\text{rec}, \textbf{x}_\text{rec})$
and $(\textbf{w}_\text{quad}, \textbf{x}_\text{quad})$ give the same kernel embedding for the RKHS given
by $ K_0$ as the latter
is a recombination of the former
(see e.g. \citep[][Eq.~14]{hayakawa2021positively}).
By subtracting, we obtain
\begin{align}
    &\lVert \mu_r - \mu_n\rVert_\H^2 \\
    &= \frac1{N^2}\left(
    \textbf{w}_\text{rec}^\top  K_1(\textbf{x}_\text{rec}, \textbf{x}_\text{rec})\textbf{w}_\text{rec} -
    2\,\textbf{w}_\text{rec}^\top  K_1(\textbf{x}_\text{rec}, \textbf{x}_\text{quad}) \textbf{w}_\text{quad}
    + \textbf{w}_\text{quad}^\top  K_1(\textbf{x}_\text{quad}, \textbf{x}_\text{quad})\textbf{w}_\text{quad}
    \right)\\
    & = \lVert \mu_r^{(1)} -
    \mu_n^{(1)}\rVert_{\H_1}^2,
\end{align}
where $\H_1$ is the RKHS given by $ K_1$
and 
\begin{equation}
    \mu_r^{(1)}:=
    \frac1N\textbf{w}_\text{rec}^\top
     K_1(\textbf{x}_\text{rec}, \cdot),
    \qquad
    \mu_n^{(1)}:=
    \frac1N\textbf{w}_\text{quad}^\top
     K_1(\textbf{x}_\text{quad}, \cdot).
\end{equation}
Now, by letting
$k_1^{1/2}(x):=\sqrt{ K(x, x)}$,
we have $\lVert K_1(x, \cdot)\rVert_{\H_1}
= k_1^{1/2}(x)$ for a point $x$.
So we have
\begin{equation}
    \lVert\mu_r^{(1)}\rVert_{\H_1}
    \le \frac1N\textbf{w}_\text{rec}^\top
    k_1^{1/2}(\textbf{x}_\text{rec}),
    \qquad
    \lVert \mu_n^{(1)}\rVert_{\H_1}\le
    \frac1N\textbf{w}_\text{quad}^\top
    k_1^{1/2}(\textbf{x}_\text{quad})
    \le \frac1N\textbf{w}_\text{rec}^\top
    k_1^{1/2}(\textbf{x}_\text{rec}),
\end{equation}
where the last inequality follows from the assumption
that $(\textbf{w}_\text{quad}, \textbf{x}_\text{quad})$ is a proper recombination of $(\textbf{w}_\text{rec}, \textbf{x}_\text{rec})$.
Therefore, we have the estimate
\begin{equation}
    \lVert \mu_r - \mu_n\rVert_\H
    = \lVert \mu_r^{(1)} - \mu_n^{(1)}\rVert_{\H_1}
    \le \lVert \mu_r^{(1)}\rVert_{\H_1}
    +\lVert\mu_n^{(1)}\rVert_{\H_1}
    \le \frac2N\textbf{w}_\text{rec}^\top
    k_1^{1/2}(\textbf{x}_\text{rec}).
\end{equation}

Finally, to prove \eqref{eq:wtp-gen},
we recall that $\textbf{x}_\text{rec}$ is an $N$-point independent sample from $g$ and $\textbf{w}_\text{rec} = \lambda(\textbf{x}_\text{rec})$,
so we obtain
\begin{align}
    \mathbb{E}_{\textbf{x}_\text{rec}}[\lVert \mu_r - \mu_n\rVert_\H]
    &\le 2\, \mathbb{E}_{\textbf{x}_\text{rec}}\!\left[
        \frac1N\textbf{w}_\text{rec}^\top k_1^{1/2}(\textbf{x}_\text{rec})
    \right]
    = 2\int \lambda(x)k_1^{1/2}(x)g(x)\dd x
    \\
    &= 
    2\int \sqrt{ K_1(x, x)}f(x)\dd x
    \le 2\left(\int  K_1(x, x) f(x)\dd x\right)^{1/2},
\end{align}
where we have used Cauchy--Schwarz in the last inequality.
\end{proof}

\subsection{Eigenvalue dacay of integral operators}
Let us consider the integral operator
\begin{equation}
    h \mapsto \int  K(\cdot, y)h(y)f(y)\dd y
\end{equation}
where 
$h \in L^2(f)
    := \{\tilde{h} \mid \text{measurable},\,
    \lVert \tilde{h}\rVert_{L^2(f)}^2
    := \int \tilde{h}(x)^2f(x)\dd x < \infty \}$,
and let $\sigma_1 \ge\sigma_2\ge\cdots \ge 0$
be eigenvalues of this operator.
This sequence of eigenvalues is known to be
closely related to the convergence rate of
kernel quadrature \citep{bach2017on}.

For the Nystr{\"o}m approximation,
we have the following estimate represented by
the eigenvalues:
\begin{thm}[\citep{hayakawa2021positively}]\label{thm:app-nys}
    For a probability density function
    $f$ on $\R^d$,
    let \rm$\textbf{x}_\text{nys}$\it\
    be an $M$-point independent sample
    from $f$.
    Let \rm$ K_0$\it\ be the rank-$(n-1)$ approximate
    kernel using \rm$\textbf{x}_\text{nys}$\it\ given by Eq. (10) in the main text.
    Then, \rm$ K_1:=  K- K_0$\it\
    satisfies
    \rm\begin{equation}
        \int  K_1(x, x)f(x)\dd x
        \le n\sigma_n + \sum_{m={n+1}}^\infty\sigma_n
        + \frac{2(n-1) K_{\max}}{\sqrt{M}}
        \left(1 + \sqrt{2\log\frac1\delta}\right)
    \end{equation}\it
    with probability at least $1-\delta$.
\end{thm}
This gives a theoretical guarantee for one step
of our algorithm, combined with Theorem \ref{thm:recall-main}.

Although the sequence of eigenvalues $\sigma_n$ 
does not have an obvious expression when $ K$ is the kernel in the middle of our algorithms BASQ,
when $ K$ is a multivariate Gaussian (RBF) kernel and
$f$ is also a Gaussian density,
we have a concrete expression of eigenvalues \citep{fas12}.

Indeed, if $ K(x, y) =
\exp(-\epsilon^2\lvert x - y\rvert^2)$
and $f(x)\propto \exp(-\alpha^2\lvert x\rvert^2)$,
in the case $d=1$,
we have $\sigma_n = ab^n$ for some constants $a>0$
and $0<b<1$ depending on $\epsilon$ and $\alpha$.
Thus, for the $d$-dimensional case, we can roughly
estimate that
$\sigma_n \le a^db^{m+1}$ if $n > m^d$.
So, by only using $n$,
we have
\begin{equation}
    \sigma_n\le a^db^{\lceil n^{1/d}\rceil}
    \le a^d b^{(n^{1/d})}
    = a^d\exp(-cn^{1/d}),
\end{equation}
for $c = \log(1/b)$.

\section{Model Analysis}
\subsection{Ablation study}
\subsubsection{Ablation study of sampling methods}

\begin{table}
\caption{Ablation study of sampling methods}
\label{table:ablation}
\centering
\setlength{\tabcolsep}{3pt}
\renewcommand{\arraystretch}{1.7}
\begin{small}
\begin{tabular}{cc|cc|cc|ccccc}
\hline
\multicolumn{2}{c|}{\textbf{Sampling}}                                                                          & \multicolumn{2}{c|}{\textbf{Prop. dist.}}                                                                    & \multicolumn{2}{c|}{\textbf{Alternate update}}                 & \multicolumn{5}{c}{\textbf{Performance metric}}                                                                                                                                                                                                                                                                                             \\ \hline
\begin{tabular}[c]{@{}c@{}}Unc.\\ sampl.\end{tabular} & \begin{tabular}[c]{@{}c@{}}Factor.\\ trick\end{tabular} & \begin{tabular}[c]{@{}c@{}}Linear\\ IVR\end{tabular} & \begin{tabular}[c]{@{}c@{}}Optimal\\ IVR\end{tabular} & \begin{tabular}[c]{@{}c@{}}Kernel\\ Update\end{tabular} & RCHQ & logMAE                                                   & logKL                                                     & \begin{tabular}[c]{@{}c@{}}wall\\ time (s)\end{tabular}          & \begin{tabular}[c]{@{}c@{}}logMAE\\ per time\end{tabular}         & \begin{tabular}[c]{@{}c@{}}logKL\\ per time\end{tabular}          \\ \hline
                                                      &                                                         &                                                      &                                                       & \ding{52}                                                       &      & \begin{tabular}[c]{@{}c@{}}-1.8750\\ $\pm 0.0435$\end{tabular} & \begin{tabular}[c]{@{}c@{}}-8.1366\\ $\pm0.2049$\end{tabular}  & \begin{tabular}[c]{@{}c@{}}407.42\\ $\pm10.139$\end{tabular}          & \begin{tabular}[c]{@{}c@{}}-0.0046\\ $\pm0.0002$\end{tabular}          & \begin{tabular}[c]{@{}c@{}}-0.0200\\ $\pm0.0010$\end{tabular}          \\
                                                      & \ding{52}                                                       &                                                      &                                                       & \ding{52}                                                       & \ding{52}    & \begin{tabular}[c]{@{}c@{}}-3.3310\\ $\pm0.5265$\end{tabular} & \begin{tabular}[c]{@{}c@{}}-9.5934\\ $\pm0.1697$\end{tabular}  & \begin{tabular}[c]{@{}c@{}}50.065\\ $\pm8.3153$\end{tabular}          & \begin{tabular}[c]{@{}c@{}}-0.0702\\ $\pm0.0222$\end{tabular}          & \begin{tabular}[c]{@{}c@{}}-0.1976\\ $\pm0.0362$\end{tabular}          \\
\ding{52}                                                     &                                                         & \ding{52}                                                    &                                                       & \ding{52}                                                      & \ding{52}    & \begin{tabular}[c]{@{}c@{}}-3.6743\\ $\pm0.0449$\end{tabular} & \begin{tabular}[c]{@{}c@{}}-9.6108\\ $\pm0.1363$\end{tabular}  & \begin{tabular}[c]{@{}c@{}}367.63\\ $\pm26.565$\end{tabular}          & \begin{tabular}[c]{@{}c@{}}-0.0100\\ $\pm0.0008$\end{tabular}          & \begin{tabular}[c]{@{}c@{}}-0.0263\\ $\pm0.0023$\end{tabular}          \\
\ding{52}                                                    & \ding{52}                                                       & \ding{52}                                                    &                                                       &                                                         & \ding{52}    & \begin{tabular}[c]{@{}c@{}}-1.5936\\ $\pm0.0016$\end{tabular} & \begin{tabular}[c]{@{}c@{}}-7.9967\\ $\pm0.0025$\end{tabular}  & \begin{tabular}[c]{@{}c@{}}\textbf{47.499}\\ $\pm8.1966$\end{tabular} & \begin{tabular}[c]{@{}c@{}}-0.0346\\ $\pm0.0060$\end{tabular}          & \begin{tabular}[c]{@{}c@{}}-0.1735\\ $\pm0.0300$\end{tabular}          \\
\ding{52}                                                     &                                                         &                                                      & \ding{52}                                                     & \ding{52}                                                       & \ding{52}    & \begin{tabular}[c]{@{}c@{}}-3.4379\\ $\pm0.2345$\end{tabular}          & \begin{tabular}[c]{@{}c@{}}\textbf{-9.7877}\\ ±\textbf{0.4589}\end{tabular} & \begin{tabular}[c]{@{}c@{}}810.45\\ $\pm14.468$\end{tabular}          & \begin{tabular}[c]{@{}c@{}}-0.0042\\ $\pm0.0003$\end{tabular}          & \begin{tabular}[c]{@{}c@{}}-0.0121\\ $\pm0.0008$\end{tabular}          \\
\ding{52}                                                     & \ding{52}                                                      &\ding{52}                                                    &                                                       & \ding{52}                                                       & \ding{52}    & \begin{tabular}[c]{@{}c@{}}\textbf{-4.0138}\\ ±\textbf{0.0078}\end{tabular} & \begin{tabular}[c]{@{}c@{}}-9.6222\\ ±0.17147\end{tabular}         & \begin{tabular}[c]{@{}c@{}}48.75\\ ±8.2176\end{tabular}           & \begin{tabular}[c]{@{}c@{}}\textbf{-0.0848}\\ ±\textbf{0.0144}\end{tabular} & \begin{tabular}[c]{@{}c@{}}\textbf{-0.2038}\\ ±\textbf{0.0379}\end{tabular} \\ \hline
\end{tabular}
\end{small}
\end{table}

We investigated the influence of each component using 10-dimensional Gaussian mixture likelihood. The performance is evaluated by taking the mean and standard deviation of five metrics when each model gathered 1,000 observations with $n=100$ batch size. LogMAE is the natural logarithmic MAE between the estimated integral value and true one, and the logKL is the natural logarithmic of the KL divergence between the estimated posterior and true one. Wall time is the overhead time until gathering 1,000 observations in seconds. LogMAE per time refers to the value that logMAE divided by the wall time. LogKL per time is the same.

%The compared models are as follows:
%\begin{enumerate}
%    \item Uncertainty sampling (Unc. sampl.)
%    \item Factorisation trick (Factor. trick)
%    \item Linearised IVR (Linear IVR)
%    \item Optimal IVR
%    \item Kernel update
%    \item RCHQ
%\end{enumerate}

Uncertainty sampling (Unc. sampl.) and factorisation trick (Factor. trick) refers to the technique explained in the section 4.2 in the main paper. Linearised IVR proposal distribution (Linear IVR) is the ones with Equations (8)-(10) in the main paper, whereas the optimal IVR is the square-root kernel IVR $g(x) = \sqrt{C^L_\textbf{y}(x,x)}\pi(x)$ derived from the Lemma 1. Kernel update refers to the type-II MLE to optimise the hyperparameters. RCHQ means whether or not to adopt RCHQ, if not, it means multi-start optimisation (that is, the same with batch WSABI.)

Sampling from optimal IVR of the square root is intractable, so we adopted the SMC sampling scheme. Firstly, supersamples $\textbf{X}_\text{super}$ are generated from prior $\pi(x)$, then we calculate the weights $ w_i = \sqrt{C^L_\textbf{y}(X_i, X_i)}\pi(x_i) / \pi(x_i) = \sqrt{C^L_\textbf{y}(X_i, X_i)}$, then we normalise them via $w^n_i := \frac{w_i}{\sum_i w_i}$, where $\sum_i w^n_i= 1$. At last, we resample subsamples $\textbf{X}_\text{quad}$ from the supersamples $\textbf{X}_\text{super}$ with the weights $w_i$. By removing the identical samples from $\textbf{X}_\text{quad}$, we can construct $\textbf{X}_\text{quad}$. This removal reduces the size of subsamples to approximately 100 times smaller, so we need to supersample at least 100 times larger than the size of the subsample $\textbf{X}_\text{quad}$. As the size of subsamples is already large, this SMC procedure is computationally demanding.

All components were examined by removing each. The ablation study of the sampling scheme in table \ref{table:ablation} shows that all components are essential for reducing overhead or faster convergence. Alternate update and uncertainty sampling contribute to the fast convergence, and factorisation trick and linearised proposal distribution reduce overhead with a negligible effect on convergence. 

\subsubsection{Ablation study of BQ modelling}
\begin{table}
\begin{small}
\caption{Ablation study of BQ modelling}
\label{table:modelling}
\centering
\setlength{\tabcolsep}{5pt}
\renewcommand{\arraystretch}{1.7}
\begin{tabular}{cccc|ccccc}
\hline
\multicolumn{4}{c|}{\textbf{BQ modelling}}                                                                                                                                                                              & \multicolumn{5}{c}{\textbf{Performance metric}}                                                                                                                                                                                                                                                                                                             \\ \hline
\begin{tabular}[c]{@{}c@{}}WSABI\\ -L\end{tabular} & \begin{tabular}[c]{@{}c@{}}WSABI\\ -M\end{tabular} & \begin{tabular}[c]{@{}c@{}}VBQ\\ (no warp)\end{tabular} & \begin{tabular}[c]{@{}c@{}}BBQ\\ (log warp)\end{tabular} & logMAE                                                            & logKL                                                             & \begin{tabular}[c]{@{}c@{}}wall\\ time (s)\end{tabular}         & \begin{tabular}[c]{@{}c@{}}logMAE\\ per time\end{tabular}         & \begin{tabular}[c]{@{}c@{}}logKL\\ per time\end{tabular}          \\ \hline
\ding{52}                                                  &                                                    &                                                         &                                                     & \begin{tabular}[c]{@{}c@{}}-4.0138\\ $\pm0.0078$\end{tabular}          & \begin{tabular}[c]{@{}c@{}}-9.6222\\ $\pm0.1715$\end{tabular}          & \textbf{\begin{tabular}[c]{@{}c@{}}48.75\\ \textbf{8.2176}\end{tabular}} & \textbf{\begin{tabular}[c]{@{}c@{}}-0.0848\\ \textbf{0.0144}\end{tabular}} & \textbf{\begin{tabular}[c]{@{}c@{}}-0.2038\\ \textbf{0.0379}\end{tabular}} \\
                                                   & \ding{52}                                                  &                                                         &                                                     & \textbf{\begin{tabular}[c]{@{}c@{}}-4.0418\\ \textbf{0.0532}\end{tabular}} & \textbf{\begin{tabular}[c]{@{}c@{}}-10.083\\ \textbf{0.2088}\end{tabular}} & \begin{tabular}[c]{@{}c@{}}814.13\\ $\pm19.813$\end{tabular}         & \begin{tabular}[c]{@{}c@{}}-0.0050\\ $\pm0.0002$\end{tabular}          & \begin{tabular}[c]{@{}c@{}}-0.0123\\ $\pm0.0006$\end{tabular}          \\
                                                   &                                                    & \ding{52}                                                       &                                                     & \begin{tabular}[c]{@{}c@{}}-2.5842\\ $\pm0.9978$\end{tabular}          & \begin{tabular}[c]{@{}c@{}}12.307\\ $\pm0.0158$\end{tabular}           & \begin{tabular}[c]{@{}c@{}}50.901\\ $\pm5.3634$\end{tabular}         & \begin{tabular}[c]{@{}c@{}}0.0534\\ $\pm0.0252$\end{tabular}           & \begin{tabular}[c]{@{}c@{}}0.2954\\ $\pm0.0254$\end{tabular}           \\
                                                   &                                                    &                                                         & \ding{52}                                                   & \begin{tabular}[c]{@{}c@{}}-3.1278\\ $\pm1.7428$\end{tabular}          & \begin{tabular}[c]{@{}c@{}}9.0425\\ $\pm0.3634$\end{tabular}           & \begin{tabular}[c]{@{}c@{}}54.092\\ $\pm5.4892$\end{tabular}         & \begin{tabular}[c]{@{}c@{}}0.0617\\ $\pm0.0285$\end{tabular}           & \begin{tabular}[c]{@{}c@{}}-0.2038\\ $\pm0.0379$\end{tabular}          \\ \hline
\end{tabular}
\end{small}
\end{table}

We investigated BQ modelling influence with the same procedure of the ablation study in the previous section. The compared models are WSABI-L, WSABI-M, Vanilla BQ (VBA), and log-warp BQ (BBQ). For the details of VBQ and BBQ modelling, see sections 3.3 and 3.4. With regard to the WSABI-M modelling, it has a disadvantageous formula in the mean posterior predictive distribution $m^L_\textbf{y}(x)$. The acquisition function is expressed as:
\begin{align}
    m^L_\textbf{y}(x) &:= \alpha + \frac12 \left(
    m_\textbf{y}(x)^2 + C_\textbf{y}(x,x)
    \right).
\end{align}

Then, the expectation of the WSABI-M is no more single term;
\begin{align}
    \mathbb{E}[m^L_\textbf{y}(x)] &:=
    \alpha 
    + \frac12 \mathbb{E}[m_\textbf{y}(x)^2]
    + \frac12 \mathbb{E}[C_\textbf{y}(x,x)]
\end{align}

As such, we cannot apply the factorisation trick for speed. Thus, we should adopt the SMC sampling scheme to sample from WSABI-M as the same procedure with the square root kernel IVR (Optimal IVR) explained in the previous section. This significantly slows down the computation with WSABI-M.

The ablation study result is shown in table \ref{table:modelling}. While WSABI-M achieves slightly better accuracy than WSABI-L, it also records the slowest computation. The WSABI-M intractable expression hinders to apply the quick sampling schemes we adopted. Vanilla BQ and BBQ (log warped BQ) shows larger errors. Therefore, the WSABI-L adoption is reasonable in this setting.

\subsubsection{Ablation study of kernel modelling}
\begin{table}[ht]
\centering
\begin{small}
\caption{Ablation study of kernel modelling}
\label{table:kernel}
\setlength{\tabcolsep}{5pt}
\renewcommand{\arraystretch}{1.7}
\begin{tabular}{c|ccccc}
\hline
Kenel                                                             & logMAE                                                              & logKL                                                              & \begin{tabular}[c]{@{}c@{}}wall \\ time (s)\end{tabular}           & \begin{tabular}[c]{@{}c@{}}logMAE\\ per time\end{tabular}           & \begin{tabular}[c]{@{}c@{}}logKL\\ per time\end{tabular}           \\ \hline
RBF                                                               & \begin{tabular}[c]{@{}c@{}}-2.9598\\ ± 0.3679\end{tabular}          & \begin{tabular}[c]{@{}c@{}}12.321\\ ± 0.0288\end{tabular}          & \begin{tabular}[c]{@{}c@{}}54.730\\ ± 1.2979\end{tabular}          & \begin{tabular}[c]{@{}c@{}}-0.0543\\ ± 0.0080\end{tabular}          & \begin{tabular}[c]{@{}c@{}}0.2349\\ ± 0.0048\end{tabular}          \\
Matérn32                                                          & \begin{tabular}[c]{@{}c@{}}-3.7328\\ ± 0.9608\end{tabular}          & \begin{tabular}[c]{@{}c@{}}12.312\\ ± 0.0577\end{tabular}          & \begin{tabular}[c]{@{}c@{}}53.502\\ ± 0.8661\end{tabular}          & \begin{tabular}[c]{@{}c@{}}-0.0701\\ ± 0.0191\end{tabular}          & \begin{tabular}[c]{@{}c@{}}0.2355\\ ± 0.0026\end{tabular}          \\
Matérn52                                                          & \begin{tabular}[c]{@{}c@{}}-4.3773\\ ± 1.4593\end{tabular}          & \begin{tabular}[c]{@{}c@{}}12.332\\ ± 0.0765\end{tabular}          & \begin{tabular}[c]{@{}c@{}}53.925\\ ± 0.9662\end{tabular}          & \begin{tabular}[c]{@{}c@{}}-0.0817\\ ± 0.0285\end{tabular}          & \begin{tabular}[c]{@{}c@{}}0.2341\\ ± 0.0027\end{tabular}          \\
Polynomial                                                        & \begin{tabular}[c]{@{}c@{}}-3.7828\\ ± 0.6803\end{tabular}          & \begin{tabular}[c]{@{}c@{}}12.208\\ ± 0.0673\end{tabular}          & \begin{tabular}[c]{@{}c@{}}52.281\\ ± 1.5491\end{tabular}          & \begin{tabular}[c]{@{}c@{}}-0.0728\\ ± 0.0152\end{tabular}          & \begin{tabular}[c]{@{}c@{}}0.2449\\ ± 0.0056\end{tabular}          \\
Exponential                                                       & \textbf{\begin{tabular}[c]{@{}c@{}}-4.6094\\ ± 1.3440\end{tabular}} & \begin{tabular}[c]{@{}c@{}}12.268\\ ± 0.0291\end{tabular}          & \textbf{\begin{tabular}[c]{@{}c@{}}54.891\\ ± 0.3428\end{tabular}} & \textbf{\begin{tabular}[c]{@{}c@{}}-0.0841\\ ± 0.0250\end{tabular}} & \begin{tabular}[c]{@{}c@{}}0.2252\\ ± 0.0009\end{tabular}          \\
\begin{tabular}[c]{@{}c@{}}Rational\\ quadratic\end{tabular}      & \begin{tabular}[c]{@{}c@{}}-3.2004\\ ± 0.6515\end{tabular}          & \begin{tabular}[c]{@{}c@{}}12.309\\ ± 0.0596\end{tabular}          & \begin{tabular}[c]{@{}c@{}}62.645\\ ± 1.7786\end{tabular}          & \begin{tabular}[c]{@{}c@{}}-0.0514\\ ± 0.0119\end{tabular}          & \textbf{\begin{tabular}[c]{@{}c@{}}0.2059\\ ± 0.0046\end{tabular}} \\
\begin{tabular}[c]{@{}c@{}}Exponentiated\\ quadratic\end{tabular} & \begin{tabular}[c]{@{}c@{}}-3.3361\\ ± 1.4484\end{tabular}          & \textbf{\begin{tabular}[c]{@{}c@{}}11.046\\ ± 0.4465\end{tabular}} & \begin{tabular}[c]{@{}c@{}}53.306\\ ± 0.1992\end{tabular}          & \begin{tabular}[c]{@{}c@{}}-0.0627\\ ± 0.0274\end{tabular}          & \begin{tabular}[c]{@{}c@{}}0.2072\\ ± 0.0076\end{tabular}          \\ \hline
\end{tabular}
\end{small}
\end{table}

We investigated kernel modelling influence with the same procedure of the ablation study in the previous section. The compared kernels are RBF (Radial Basis Function, as known as squared exponential, or Gaussian), Matérn32, Matérn52, Polynomial, Exponential, Rational quadratic, and exponentiated quadratic. The quadrature was performed via RCHQ with the weighted sum of the mean predictive distribution of the optimised GP. Exponential kernel marked the best accuracy in the evidence inference, whereas the KL divergence of posterior is embarrassingly erroneous. This is because all kernels examined in this section is not warped; thus, the GP-modelled likelihood is not non-negative.

\subsection{Hyperparameter sensitivity analysis}
\subsubsection{Analysis results}
\begin{table}
\caption{Sensitivity analysis with functional ANOVA}
\label{table:anova}
\centering
\begin{tabular}{llllll}
\toprule
\begin{tabular}[c]{@{}l@{}}hyperparameters\end{tabular} & logMAE          & logKL           & \begin{tabular}[c]{@{}l@{}}wall\\ time\end{tabular}            & \begin{tabular}[c]{@{}l@{}}logMAE\\ per time\end{tabular} & \begin{tabular}[c]{@{}l@{}}logKL \\ per time\end{tabular} \\
\midrule
N                                                           & 0.0835          & 0.0465          & \textbf{0.6347}          & 0.1729                                                   & 0.1811                                                    \\
M                                                           & 0.1041          & 0.0824          & 0.2497          & \textbf{0.6401}                                           & \textbf{0.6789}                                           \\
r                                                           & 0.0041          & 0.0226          & 0.0071          & 0.0040                                                   & 0.0024                                                    \\
N,M                                                         & 0.0710          & 0.1134          & 0.0903 & 0.1077                                                    & 0.1206                                                    \\
N,r                                                         & 0.1021          & 0.1299          & 0.0058          & 0.0062                                                    & 0.0024                                                    \\
M,r                                                         & \textbf{0.4598}          & \textbf{0.4381} & 0.0059          & 0.0604                                                    & 0.0123                                                    \\
N,M,r                                                       & 0.1754 & 0.3116          & 0.1671          & 0.0065                                                    & 0.0023\\
\bottomrule
\end{tabular}
\end{table}

\begin{figure}
  \centering
  \includegraphics[width=0.7\textwidth,center]{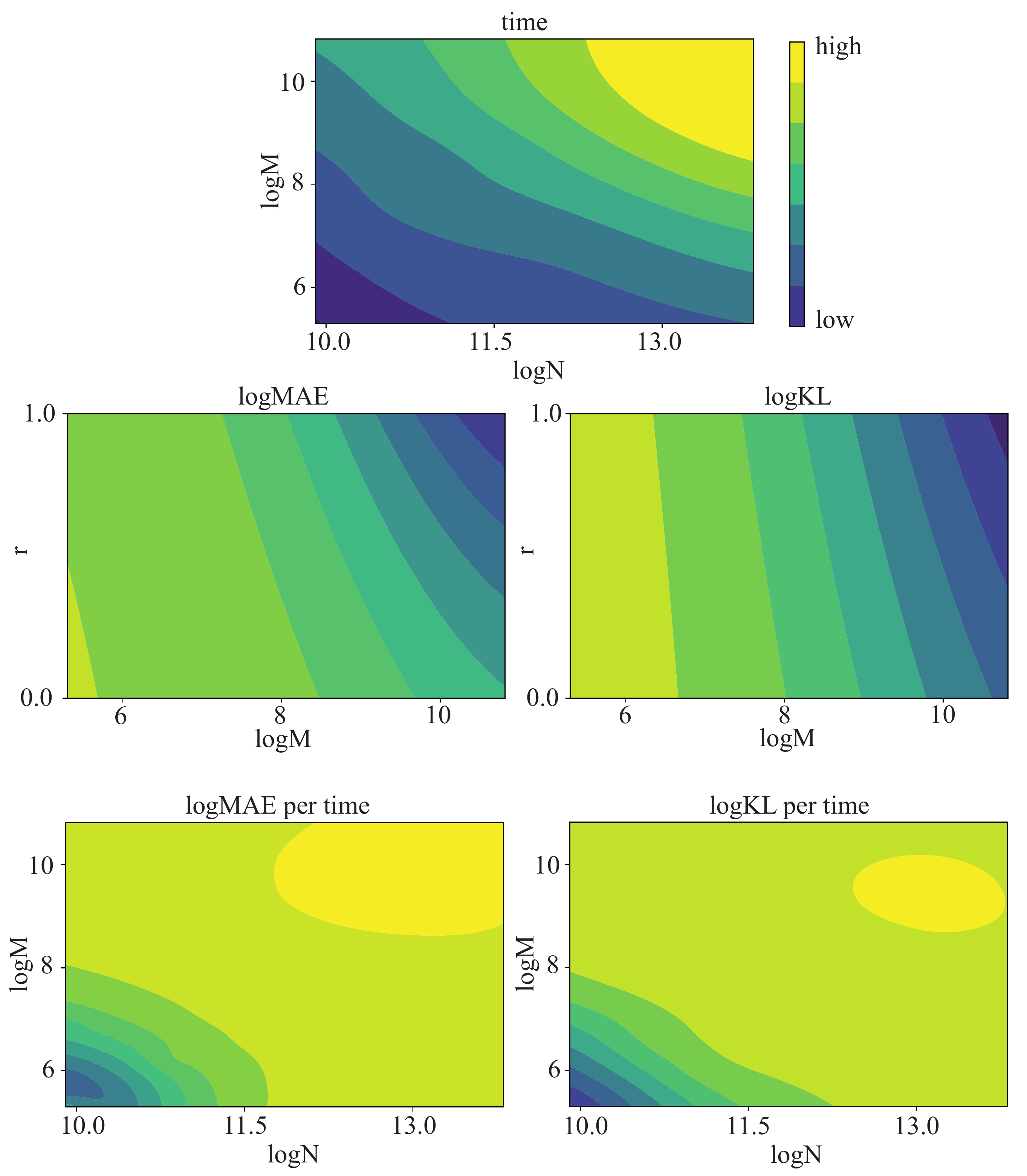}
  \caption{Sensitivity analysis of the hyperparameters over the metrics} \label{fig:anova}
\end{figure}

The hyperparameter sensitivity analysis is performed in the same setting as the previous section that adopts a ten-dimensional Gaussian mixture. The analysis was performed with functional Analysis of Variance (ANOVA) \citep{hutter2014efficient}. The functional ANOVA is the statistical method to decompose the variance $V$ of a black box function $f$ into additive components $VU$ associated with each subset of hyperparameters. \citep{hutter2014efficient} adopts random forest for efficient decomposition and marginal prediction over each hyperparameter. The hyperparameters to be analysed are the number of subsamples for recombination $N$, the number of samples for the Nyström method, and the partition ratio in the IVR proposal distribution $r$. They need to satisfy the following relationship; $N \gg M > n$, where $n$ is the batch size. We typically take $n=100$, so $M$ should be larger than at least 200, and $N$ should be larger than at least 20,000. Grid search was adopted for the hyperparameter space, with the range of $N$ = {20,000, 50,000, 100,000, 500,000, 1,000,000}, $M$ = {200, 500, 1,000, 5,000, 10,000}, and $r$ = {0.0, 0.25, 0.5, 0.75, 1.0}, resulting in 125 data points. To compensate for the dynamic range difference, a natural logarithmic $\log N$ and $\log M$ were used as the input.

The result is shown in Table \ref{table:anova}. Each value refers to the fraction of the decomposed variance, corresponding to the importance of each hyperparameter over the performance metric. Functional ANOVA evaluates the main effect and the pairwise interaction effect. Figure \ref{fig:anova} shows the marginal predictions on the important two hyperparameters for each performance metric.

As the most obvious case, we will look into the wall time. The most important hyperparameter was $N$, and the second was $M$. This is well correlated to the theoretical aspect. The overhead computation time of BASQ can be decomposed into two components; subsampling and RCHQ. The $N$ subsampling is dependent on $N$ as $\mathcal{O}(n^2/2 + n_\text{comp} N)$. $n_\text{comp}$ is way less than $M$ or $n^2$ and is insensitive to the hyperparameter variation or GP updates as we designed it to be sparse. The SMC sampling for $M$ is negligible as $\mathcal{O}(N + M)$. The RCHQ is $\mathcal{O}(NM + M^2 \log n +Mn^2 \log(N/n))$. Comparing the complexity, the RCHQ stands out. Therefore, the whole BASQ algorithm complexity is dominated by RCHQ. While $M$ has the squared component, $N$ itself is as large as $M^2$. Therefore, the selected two hyperparameters $N$ and $M$ align with the theory. Figure \ref{fig:anova} agrees the above analysis.

The logMAE and logKL metrics have a similar trend to each other. In fact, their correlation coefficient was 0.6494. This makes sense because both metrics are determined by the functional approximation accuracy of likelihood $\ell(x)$. While increasing $M$ is always beneficial in any $r$, varying $r$ is effective in larger $M$. At last, the metrics of performance per wall time is the combination of these effects. Obviously, time is the dominant factor, so we should limit the $N$ and $M$ as small as possible. The most important hyperparameter was $M$. This is a natural consequence because $M$ affects the overhead increment less than $N$ but contributes to reducing errors more.

\subsubsection{A guideline to select hyperparameters}
The main takeaway from the functional ANOVA analysis is that the accuracy with and without the time constraint has an opposite trend. Therefore, the best hyperparameter set is dependent on the overhead time allowance. The expensiveness of likelihood evaluation determines this. Per the likelihood query time per iteration, we should increase $M$ and $N$ for faster convergence.

In the cheap likelihood case, the most relevant metric is logMAE per time and logKL per time. As we should minimise the time, we choose the minimal size for $N$ and $M$ to minimise the overhead. As the typical hyperparameter set is $(n, N, M, r) = (100, 200, 20,000, 0.5)$, and these are the minimum values for $N$ and $M$ at given $n$. The remained choice is the selection of $r$. As shown in Figure \ref{fig:anova}, $r=1$, namely, UB proposal distribution, was the best selection in the Gaussian mixture likelihood case. A similar trend is observed in the synthetic dataset. However, as we observed in the real-world dataset cases, some likelihoods showed that $r=0.5$, namely IVR proposal distribution outperformed the UB. Therefore, the $r=1$ might be the first choice., and $r=0.5$ is the second choice.

In the expensive likelihood case, we can increase the $N$ and $M$ because the overhead time is less significant than the likelihood query time. The logMAE and logKL without time constraints are good guidelines for tuning the hyperparameters. As $M$ is the most significant hyperparameter, we wish to increase $M$ first. However, we have to increase $N$ under the constraint $N \gg M$, necessary for RCHQ fast convergence.
Empirically, we recommend increasing the $M$ three times larger than $N$ from the minimum set because the importance factor ratio in Table \ref{table:anova} is roughly three times. And the increment of $M$ should be corresponded to the likelihood query time $t_\text{likelihood}$ over the RCHQ computation time $t_\text{RCHQ}$. We increase the $M$ in accordance with the ratio $r_\text{comp} := t_\text{likelihood} / t_\text{RCHQ}$. Thus, the $M$ and $N$ is determined as $M = r_\text{comp} M_\text{min}$, $N = \frac{r_\text{comp}}{3} N_\text{min}$, where $M_\text{min} = 200, N_\text{min}=20,000$.

\section{Analytical form of integrals}
\subsection{Gaussian identities}
\begin{align}
    \mathcal{N}(\textbf{x}; \textbf{m}_1, \boldsymbol\Sigma_1)\mathcal{N}(\textbf{x}; \textbf{m}_2, \boldsymbol\Sigma_2) 
    &= C_c\mathcal{N}(\textbf{x}; \textbf{m}_c, \boldsymbol\Sigma_c)\\
    \int \mathcal{N}(\textbf{x}; \textbf{m}_1, \boldsymbol\Sigma_1)\mathcal{N}(\textbf{x}; \textbf{m}_2, \boldsymbol\Sigma_2) d\textbf{x}
    &= C_c\\
    \mathcal{N}(\textbf{Ax+b}; \textbf{m}, \boldsymbol\Sigma)
    &= \sqrt{\frac{|2\pi(\textbf{A}^\top \boldsymbol\Sigma^{-1} \textbf{A})^{-1}|}{|2\pi\boldsymbol\Sigma|}}
    \mathcal{N}\left(
    \textbf{x}; \textbf{A}^{-1} \textbf{m} - \textbf{b}, (\textbf{A}^\top \boldsymbol\Sigma^{-1} \textbf{A})^{-1}
    \right)
\end{align}

where
\begin{align}
    \boldsymbol\Sigma_c &= \left[ \boldsymbol\Sigma^{-1}_{1} + \boldsymbol\Sigma^{-1}_{2} \right]^{-1}\\
    \textbf{m}_c &= \left[ \boldsymbol\Sigma^{-1}_{1} + \boldsymbol\Sigma^{-1}_{2} \right]^{-1} \left( \boldsymbol\Sigma^{-1}_1 \textbf{m}_1 + \boldsymbol\Sigma^{-1}_2 \textbf{m}_2 \right)\\
    C_c &= \mathcal{N}(\textbf{m}_1; \textbf{m}_2, \boldsymbol\Sigma_1 + \boldsymbol\Sigma_2)
\end{align}

In the finite number of product case:
\begin{align}
    \prod^n_{i=1} \mathcal{N}(\textbf{x}; \textbf{m}_i, \boldsymbol\Sigma_i) 
    &= C_m \mathcal{N}(\textbf{x}; \textbf{m}_m, \boldsymbol\Sigma_m)\\
    \int \prod^n_{i=1} \mathcal{N}(\textbf{x}; \textbf{m}_i, \boldsymbol\Sigma_i) d\textbf{x}
    &= C_m \int \mathcal{N}(\textbf{x}; \textbf{m}_m, \boldsymbol\Sigma_m) d\textbf{x}\\
    &= C_m
\end{align}

where
\begin{align}
    \boldsymbol\Sigma_m &= \left[ \sum^n_{i=1} \left( \boldsymbol\Sigma^{-1}_{i} \right) \right]^{-1}\\
    \textbf{m}_m &= \left[ \sum^n_{i=1} \left( \boldsymbol\Sigma^{-1}_{i} \right) \right]^{-1} \sum^n_{i=1} \left( \boldsymbol\Sigma^{-1}_i \textbf{m}_i \right)\\
    C_m &= \exp \left[
    -\frac{1}{2}
    \left\{
    (n-1)d \log 2\pi - \sum^n_{i=1} \log \left| \boldsymbol\Sigma_i^{-1}\right|
    + \log \left| \sum^n_{i=1} \left( \boldsymbol\Sigma^{-1}_{i} \right) \right| \right. \right.\\
    & \qquad \quad \left. \left. + \sum^n_{i=1} \left( (\boldsymbol\Sigma_i^{-1} \textbf{m}_i)^\top \boldsymbol\Sigma_i (\boldsymbol\Sigma_i^{-1} \textbf{m}_i) \right)
    - \left( \sum^n_{i=1} \left( \boldsymbol\Sigma_i^{-1} \textbf{m}_i \right) \right)^{\top}
    \left( \sum^n_{i=1} \left( \boldsymbol\Sigma^{-1}_{i} \right) \right)^{-1}
    \left( \sum^n_{i=1} \left( \boldsymbol\Sigma_i^{-1} \textbf{m}_i \right) \right)
    \right\}
    \right]
\end{align}

\subsection{Definitions}
$x_*$: predictive data points, $x_* \in \mathbb{R}^d$\\
$\textbf{X}$: the observed data points, $\textbf{X} \in \mathbb{R}^{n \times d}$\\
$\textbf{y} = \sqrt{2 (\ell_{\text{true}}(\textbf{X}) - \alpha)}$: the (warped) observed likelihood, $\textbf{y} \in \mathbb{R}^n$\\
$\pi(x_*) = \mathcal{N}(x_*; \mu_{\pi}, \boldsymbol\Sigma_{\pi})$: prior distribution,\\
$v^{'}$: a kernel variance,\\
$l$: a kernel lengthscale,\\
$ K(x_*, \textbf{X}) = v \mathcal{N}(x_*; \textbf{X}, \textbf{W})$: a RBF kernel,\\
$v = v^{'} \sqrt{|2\pi \textbf{W}|}$: a normalised kernel variance,\\
$\textbf{W}$: a diagonal covariance matrix whose diagonal elements are the lengthscales of each dimension,\\
$\textbf{W} = \begin{bmatrix}
l^2 & \textbf{0} \\
\textbf{0} & l^2
\end{bmatrix}$\\
$\textbf{I}$: The identity matrix,\\
$ \textbf{K}_{XX} =  K(\textbf{X}, \textbf{X})$: a kernel over the observed data points.

\subsection{Warped GPs as Gaussian Mixture}
\subsubsection{Mean}

\begin{align}
    m^L_\textbf{y}(x_*) &= \alpha + \frac{1}{2} \tilde m_{\textbf{y}}(x_*)^2 \\
    &= \alpha + \frac{1}{2} \textbf{y}^{T}  \textbf{K}^{-1}_{XX}  K(\textbf{X}, x_*)  K(x_*, \textbf{X})   \textbf{K}^{-1}_{XX}\textbf{y}\\
    &= \alpha + \frac{1}{2} \sum_{i,j} \boldsymbol\omega_{i}\boldsymbol\omega_{j} K(X_{i}, x_*) K(x_*, X_{j})\\
    &= \alpha + \sum_{i,j} w^{m}_{ij}\mathcal{N} \left(x_*; \frac{X_i + X_j}{2}, \frac{\textbf{W}}{2} \right)
\end{align}

where\\
Woodbury vector: \quad
$\boldsymbol\omega =  \textbf{K}^{-1}_{XX}\textbf{y}$, \\
mean weight: \quad
$w^{m}_{ij} = \frac{1}{2} v^2 \boldsymbol\omega_i \boldsymbol\omega_j \mathcal{N}(X_i; X_j, 2\textbf{W})$
\subsubsection{Variance}
\begin{align}
\begin{split}
C_{\textbf{y}}^{L}(x_*, x^{'}_*) &= \tilde m_{\textbf{y}}(x) \tilde C_{\textbf{y}}(x, x') \tilde m_{\textbf{y}}(x')\\
&= \left[  K(x_*, \textbf{X})\boldsymbol\omega \right]^\top
\left[  K(x_*, x^{'}_*)
-  K(x_*, \textbf{X}) \textbf{K}_{XX}^{-1} K(\textbf{X}, x^{'}_*) \right]
\left[  K(x^{'}_*, \textbf{X})\boldsymbol\omega \right]\\
&= \boldsymbol\omega^{\top}  K(\textbf{X}, x_*)  K(x_*, x^{'}_*) K(x^{'}_*, X)\boldsymbol\omega\\
& \quad - \boldsymbol\omega^{\top}  K(\textbf{X}, x_*)  K(x_*, \textbf{X}) \textbf{K}_{XX}^{-1} K(\textbf{X}, x^{'}_*) K(x^{'}_*, \textbf{X})\boldsymbol\omega\\
&=  \sum_{i,j} \boldsymbol\omega_{i}\boldsymbol\omega_{j} K(X_i, x_*)  K(x_*, x^{'}_*) K(x^{'}_*, X_j)\\
& \quad - \sum_{i,j} \boldsymbol\omega_{i}\boldsymbol\omega_{j} \sum_{k,l} \boldsymbol\Omega_{kl}  K(X_j, x_*)  K(x_*, X_i) K(X_k, x^{'}_*) K(x^{'}_*, X_l)\\
&=  \sum_{i,j} w^{v}_{ij}
\mathfrak{C_{v}}(i,j)
- \sum_{i,j} \sum_{k,l} w^{vv}_{ijkl} \mathfrak{C_{vv}}(i,j,k,l)
\end{split}
\end{align}

where\\
The inverse kernel weight: \quad
$\boldsymbol\Omega_{kl} =  \textbf{K}_{XX}^{-1}(k, l)$\\
the first variance weight: \quad
$w^{v}_{ij} = v^3 \boldsymbol\omega_{i}\boldsymbol\omega_{j}$\\
the second variance weight: \quad
$w^{vv}_{ijkl} = v^4 \boldsymbol\omega_{i}\boldsymbol\omega_{j}\boldsymbol\Omega_{kl}$\\
the first variance Gaussian variable

$\mathfrak{C_{v}}(i,j) =
\mathcal{N}
\left(
\begin{bmatrix}
x_* \\
x^{'}_*\\
x^{'}_*\\
\end{bmatrix}
;
\begin{bmatrix}
X_i \\
X_j \\
x_*
\end{bmatrix}
,
\begin{bmatrix}
\textbf{W} & \textbf{0} & \textbf{0}\\
\textbf{0} & \textbf{W} & \textbf{0}\\
\textbf{0} & \textbf{0} & \textbf{W}
\end{bmatrix}
\right)$\\

the second variance Gaussian variable

$\mathfrak{C_{vv}}(i,j,k,l) =
\mathcal{N}
\left(
\begin{bmatrix}
x_* \\
x_*\\
x^{'}_* \\
x^{'}_*
\end{bmatrix}
;
\begin{bmatrix}
X_i \\
X_j \\
X_k \\
X_l
\end{bmatrix}
,
\begin{bmatrix}
\textbf{W} & \textbf{0} & \textbf{0} & \textbf{0}\\
\textbf{0} & \textbf{W} & \textbf{0} & \textbf{0}\\
\textbf{0} & \textbf{0} & \textbf{W} & \textbf{0}\\
\textbf{0} & \textbf{0} & \textbf{0} & \textbf{W} 
\end{bmatrix}
\right)$

\newpage
\subsubsection{Symmetric variance}
Here we consider $x_* = x^\prime_*$ and $x_*$ is a sample with dimension $d$:

\begin{align}
\begin{split}
C_{\textbf{y}}^{L}(x_*, x_*) &= C_{\textbf{y}}^{L}(x_*)\\
&= \boldsymbol\omega^{\top}  K(\textbf{X}, x_*)  K(x_*, x_*) K(x_*, \textbf{X})\boldsymbol\omega\\
& \quad - \boldsymbol\omega^{\top}  K(\textbf{X}, x_*)  K(x_*, \textbf{X}) \textbf{K}_{XX}^{-1} K(\textbf{X}, x_*) K(x_*, \textbf{X})\boldsymbol\omega\\
&= \sum_{i,j} \boldsymbol\omega_{i}\boldsymbol\omega_{j} K(X_i, x_*)  K(x_*, x_*) K(x_*, X_j)\\
& \quad - \sum_{i,j} \boldsymbol\omega_{i}\boldsymbol\omega_{j} \sum_{k,l} \boldsymbol\Omega_{kl}  K(X_j, x_*)  K(x_*, X_i)  K(X_k, x_*) K(x_*, X_l)\\
& = \frac{v}{\sqrt{|2\pi \textbf{W}|}} \sum_{i,j} \boldsymbol\omega_{i}\boldsymbol\omega_{j}  K(X_i, x_*) K(x_*, X_j)\\
& \quad - \sum_{i,j} \boldsymbol\omega_{i}\boldsymbol\omega_{j} \sum_{k,l} \boldsymbol\Omega_{kl}  K(X_j, x_*)  K(x_*, X_i)  K(X_k, x_*) K(x_*, X_l)\\
& = \frac{v^3}{\sqrt{|2\pi \textbf{W}|}} \sum_{i,j} \boldsymbol\omega_{i}\boldsymbol\omega_{j} \mathcal{N}(X_i; X_j, 2\textbf{W})  \mathcal{N} \left(x_*; \frac{X_i + X_j}{2}, \frac{\textbf{W}}{2} \right)\\
& \quad - v^4 \sum_{i,j} \left[ \boldsymbol\omega_{i}\boldsymbol\omega_{j} \sum_{k,l} \left\{ \boldsymbol\Omega_{kl} \mathcal{N}(X_l; X_i, 2\textbf{W}) \mathcal{N} \left(x_*; \frac{X_l + X_i}{2}, \frac{\textbf{W}}{2} \right) \right. \right.\\
& \left. \left. \qquad \qquad \qquad \qquad \qquad \cdot \mathcal{N}(X_k; X_j, 2\textbf{W}) \mathcal{N} \left(x_*; \frac{X_k + X_j}{2}, \frac{\textbf{W}}{2} \right) \right\} \right]\\
& = \sum_{i,j} w^{'v}_{ij} \mathcal{N} \left(x_*; \frac{X_i + X_j}{2}, \frac{\textbf{W}}{2} \right) - \sum_{i,j} \sum_{k,l} w^{'vv}_{ijkl} \mathcal{N} \left(x_*; \frac{X_i + X_j + X_k + X_l}{4}, \frac{\textbf{W}}{4} \right)
\end{split}\label{eq:variance}
\end{align}

where\\
$w^{'v}_{ij} = \frac{h^3}{\sqrt{|2\pi \textbf{W}|}} \boldsymbol\omega_{i}\boldsymbol\omega_{j} \mathcal{N}(X_i; X_j, 2\textbf{W})$\\
$w^{'vv}_{ijkl} = h^4 \boldsymbol\omega_{i}\boldsymbol\omega_{j}\boldsymbol\Omega_{kl} \mathcal{N}(X_l; X_i, 2\textbf{W}) \mathcal{N}(X_k; X_j, 2\textbf{W}) 
\mathcal{N} \left( \frac{X_k + X_j}{2}; \frac{X_i + X_l}{2}, \textbf{W} \right)$

\subsection{Model evidence}
The distribution over integral $Z$ is given by:

\begin{align}
    p(Z|\textbf{y}) &= \int p\big(Z|\ell(x_*)\big)p\big(\ell(x_*)|\textbf{y}\big) dx\\
    &= p\big(Z|\ell(x_*)\big)\mathcal{N}\big(\ell(x_*); m^L_\textbf{y}(x_*), C^L_\textbf{y}(x_*)\big)\\
    &= \mathcal{N}\Big(Z; \mathbb{E}\big[ Z | \textbf{y}\big], \text{var}\big[Z | \textbf{y}\big] \Big)
\end{align}

\subsubsection{Mean of the integral}
\begin{align}
    \mathbb{E}\big[ Z | \textbf{y}\big] &= \mathbb{E}\big[ m^L_\text{y}\big]\\
    &= \int m^L_\textbf{y}(x_*)\pi(x_*)dx_*\\
    &= \alpha + \frac{1}{2} \int \tilde m_{\textbf{y}}^2(x_*)\pi(x_*)dx_*\\
    &= \alpha +
    \sum_{i,j} w^{m}_{ij} \int \mathcal{N} \left(x_*; \frac{X_i + X_j}{2}, \frac{\textbf{W}}{2} \right)
    \mathcal{N}(x_*; \mu_{\pi}, \boldsymbol\Sigma_\pi) dx_*\\
    &= \alpha + \sum_{i,j} w^m_{ij} \mathcal{N} \left(\frac{X_i + X_j}{2}; \mu_\pi, \frac{\textbf{W}}{2} + \boldsymbol\Sigma_\pi \right)
\end{align}

\subsubsection{Variance of the integral}
\begin{align}
\begin{split}
    \text{var}\big[ Z | \textbf{y}\big] &= \text{var}\big[C^L_\textbf{y}\big]\\
    &= \iint \pi(x_*) C^L_\textbf{y}(x_*, x^{'}_*) \pi(x^{'}_*) dx_* dx^{'}_*\\
    &= \iint \left( \boldsymbol\omega^\top K(\textbf{X}, x_*) K(x_*, x^{'}_*)
    K(x^{'}_*, \textbf{X})\boldsymbol\omega \pi(x_*)\pi(x^{'}_*) \right.\\
    & \quad \left. - \boldsymbol\omega^\top K(\textbf{X}, x_*) K(x_*, \textbf{X})
    \textbf{K}_{XX}^{-1} K(\textbf{X}, x^{'}_*) K(x^{'}_*, \textbf{X})\boldsymbol\omega \pi(x_*)\pi(x^{'}_*) \right) dx_* dx^{'}_*\\
    &= \sum_{i,j} \boldsymbol\omega_i \boldsymbol\omega_j
    \iint K(X_i, x_*) K(x_*, x^{'}_*) K(x^{'}_*, X_j) \pi(x_*)\pi(x^{'}_*) dx_* dx^{'}_*\\
    &\quad - \sum_{i,j} \boldsymbol\omega_i \boldsymbol\omega_j \sum_{k,l} \boldsymbol\Omega_{kl}
    \iint K(X_j, x_*) K(x_*, X_i) K(X_k, x^{'}_*) K(x^{'}_*, X_l) \pi(x_*)\pi(x^{'}_*) dx_* dx^{'}_*\\
    &= \sum_{i,j} \boldsymbol\omega_i \boldsymbol\omega_j h^3
    \int \left[ \mathcal{N}(x^{'}_*; \mu_\pi, \Sigma_\pi) \mathcal{N}(x^{'}_*; X_j, \textbf{W})
    \int \mathcal{N}(x_*; X_i, \textbf{W}) \mathcal{N}(x_*; x^{'}_*, \textbf{W})
    \mathcal{N}(x_*; \mu_\pi, \Sigma_\pi) dx_* \right] dx^{'}_*\\
    &\quad - \sum_{i,j} \sum{k,l} \boldsymbol\omega_i \boldsymbol\omega_j \boldsymbol\Omega_{kl} h^4
    \int \mathcal{N}(x_*; X_j, \textbf{W}) \mathcal{N}(x_*; X_i, \textbf{W}) \mathcal{N}(x_*; \mu_\pi, \Sigma_\pi) dx_*\\
    & \hspace{80pt} \cdot \int \mathcal{N}(x^{'}_*; X_k, \textbf{W}) \mathcal{N}(x^{'}_*; x_l, \textbf{W}) \mathcal{N}(x^{'}_*; \mu_\pi, \Sigma_\pi) dx^{'}_*\\
    &= \sum_{i,j} \boldsymbol\omega_i \boldsymbol\omega_j h^3
    \int \mathcal{N}(x^{'}_*; \mu_\pi, \Sigma_\pi) \mathcal{N}(x^{'}_*; X_j, \textbf{W})
    \mathcal{N}(x^{'}_*; X_i, \textbf{W}) \mathcal{N} \left(\frac{X_i + x^{'}_*}{2};
    \mu_\pi, \frac{\textbf{W}}{2} + \boldsymbol\Sigma_\pi \right) dx^{'}_*\\
    &\quad - \sum_{i,j} \sum{k,l} \boldsymbol\omega_i \boldsymbol\omega_j
    \boldsymbol\Omega_{kl} h^4 \left[ \mathcal{N}(X_j; X_i, 2W) \mathcal{N}
    \left(\frac{X_i+X_j}{2}; \mu_\pi, \frac{\textbf{W}}{2}+\boldsymbol\Sigma_\pi \right) \right]\\
    & \hspace{80pt} \cdot \left[ \mathcal{N}(X_k; x_l, 2W) \mathcal{N}
    \left(\frac{X_k+X_l}{2}; \mu_\pi, \frac{\textbf{W}}{2}+\boldsymbol\Sigma_\pi \right) \right]\\
    &= \sum_{i,j} \boldsymbol\omega_i \boldsymbol\omega_j h^3
    \int \mathcal{N}(x^{'}_*; \mu_\pi, \boldsymbol\Sigma_\pi) \mathcal{N}(x^{'}_*; X_i, \textbf{W}) \mathcal{N}(x^{'}_*; X_j, \textbf{W}) 2^d \mathcal{N} \left( x^{'}_*; 2\mu_\pi - X_i/2, 2\textbf{W} + 4\boldsymbol\Sigma_\pi \right) dx^{'}_*\\
    &\quad - \sum_{i,j}\sum_{k,l} w^{vv}_{ijkl} \mathfrak{K}_{vv}(i,j,k,l)\\
    &= \sum_{i,j} 2^d w^v_{ij} \mathfrak{K}_v(i,j) - \sum_{i,j}\sum_{k,l}w^{vv}_{ijkl} \mathfrak{K}_{vv}(i,j,k,l)\\
\end{split}
\end{align}

where
\begin{align}
    \mathfrak{K_{v}}(i,j)
    &= \mathcal{N}(X_i; X_j, 2\textbf{W}) \mathcal{N} \left(\frac{X_i+X_j}{2}; \mu_{\pi}, \frac{\textbf{W}}{2}+\boldsymbol\Sigma_{\pi} \right)\\
    &\mathcal{N} \left[
    \left( 2\textbf{W}^{-1} + \boldsymbol\Sigma^{-1}_{\pi} \right)^{-1} \left( \textbf{W}^{-1}(X_i + X_j) + \boldsymbol\Sigma^{-1}_{\pi} \mu_{\pi} \right);
    2\mu_{\pi} - \frac{X_i}{2},
    \left( 2\textbf{W}^{-1} + \boldsymbol\Sigma^{-1}_{\pi} \right)^{-1} + 2\textbf{W} + 2\boldsymbol\Sigma_{\pi}
    \right]\\
    \mathfrak{K_{vv}}(i,j,k,l) &= \left[ \mathcal{N}(X_j; X_i, 2\textbf{W}) \mathcal{N} \left(\frac{X_i+X_j}{2}; \mu_{\pi}, \frac{\textbf{W}}{2}+\boldsymbol\Sigma_{\pi} \right) \right]
    \left[ \mathcal{N}(X_k; X_l, 2\textbf{W}) \mathcal{N} \left(\frac{X_k+X_l}{2}; \mu_{\pi}, \frac{\textbf{\textbf{W}}}{2}+\boldsymbol\Sigma_{\pi} \right) \right]
\end{align}

\subsection{Posterior inference}
\subsubsection{Joint posterior}
\begin{align}
    p(x) = \frac{m^L_\textbf{y}(x)\pi(x)}{\mathbb{E}[Z|\textbf{y}]}
\end{align}

\subsubsection{Marginal posterior}
The marginal posterior can be on obtained from Gaussian mixture form of joint posterior. Thanks to the Gaussianity, marginal posterior can be easily derived by extracting the d-th element of matrices in the following mixture of Gaussians.

\begin{align}
  p(x) &= \frac{\alpha}{\mathbb{E}[Z|\textbf{y}]} +
  \sum_{i,j} w^{p}_{ij} \mathcal{N}(x_*; \mu_p, \boldsymbol\Sigma_p)
\end{align}

where\\
$w^p_{ij} =  \frac{w^{m}_{ij}}{\mathbb{E}[Z|\textbf{y}]} \mathcal{N} \left(\frac{X_i + X_j}{2}; \mu_{\pi}, \frac{\textbf{W}}{2} + \boldsymbol\Sigma_{\pi} \right) $\\
$\boldsymbol\Sigma_p = (2\textbf{W}^{-1} + \boldsymbol\Sigma_{\pi}^{-1})^{-1}$\\
$\mu_p = \boldsymbol\Sigma_p (\textbf{W}^{-1}(X_i + X_j) + \boldsymbol\Sigma_{\pi}^{-1} \mu_{\pi})$

\subsubsection{Conditional posterior}
The conditional posterior $p \left( x; d=d \,|\, d=\textbf{D} \setminus \textbf{D}(\geq d) \right)$ can be derived from the Gaussian mixture form of joint posterior. We can obtain the conditional posterior via applying the following relationship to each Gaussian:
Assume $\textbf{x} \sim \mathcal{N}(\textbf{x}; \boldsymbol\mu, \boldsymbol\Sigma)$ where
\begin{align}
    \textbf{x} = 
    \begin{bmatrix}
        \textbf{x}_a \\
        \textbf{x}_b\\
    \end{bmatrix}
    \qquad
    \boldsymbol\mu = 
    \begin{bmatrix}
        \boldsymbol\mu_a \\
        \boldsymbol\mu_b\\
    \end{bmatrix}
    \qquad
    \boldsymbol\Sigma = 
    \begin{bmatrix}
        \boldsymbol\Sigma_a & \boldsymbol\Sigma_c\\
        \boldsymbol\Sigma_c^\top & \boldsymbol\Sigma_b\\
    \end{bmatrix}
\end{align}

Then
\begin{align}
    p(\textbf{x}_a) | p(\textbf{x}_b) = 
    \mathcal{N}(\textbf{x}_a; \hat{\boldsymbol\mu}_a, \hat{\boldsymbol\Sigma}_a)
    \qquad
    \begin{cases}
        \hat{\boldsymbol\mu}_a = \boldsymbol\mu_a + \boldsymbol\Sigma_c\boldsymbol\Sigma_b^{-1}(\textbf{x}_b - \boldsymbol\mu_b)\\
        \hat{\boldsymbol\Sigma}_a = \boldsymbol\Sigma_a - \boldsymbol\Sigma_c\boldsymbol\Sigma_b^{-1}\boldsymbol\Sigma_c^\top
    \end{cases}\\
    p(\textbf{x}_b) | p(\textbf{x}_a) = 
    \mathcal{N}(\textbf{x}_b; \hat{\boldsymbol\mu}_b, \hat{\boldsymbol\Sigma}_b)
    \qquad
    \begin{cases}
        \hat{\boldsymbol\mu}_b = \boldsymbol\mu_b + \boldsymbol\Sigma_c\boldsymbol\Sigma_a^{-1}(\textbf{x}_a - \boldsymbol\mu_a)\\
        \hat{\boldsymbol\Sigma}_b = \boldsymbol\Sigma_b - \boldsymbol\Sigma_c\boldsymbol\Sigma_a^{-1}\boldsymbol\Sigma_c^\top
    \end{cases}
\end{align}

\section{Uncertainty sampling}
\subsection{Analytical form of acquisiton function}
\subsubsection{Acquisiton function as Gaussian Mixture}
We set the acquisition function $A(x)$ as the product of the variance and the prior. As is shown in Eq. \eqref{eq:variance}, When we provide the predictive samples $x_*$:

\begin{align}
A(x_*) &= C_{\textbf{y}}^{L}(x_*, x_*) \pi(x_*)\\
&= C_{\textbf{y}}^{L}(x_*) \pi(x_*)\\
&= \left( \sum_{i,j} w^{'v}_{ij} \mathcal{N} \left(x_*; \frac{X_i + X_j}{2}, \frac{\textbf{W}}{2} \right) - \sum_{i,j} \sum_{k,l} w^{'vv}_{ijkl} \mathcal{N} \left(x_*; \frac{X_i + X_j + X_k + X_l}{4}, \frac{\textbf{W}}{4} \right)
\right)\\
& \quad \cdot \mathcal{N} \left(x_*; \mu_{\pi}, \boldsymbol\Sigma_{\pi} \right)\\
&= \sum_{i,j} w^{'A}_{ij} \mathcal{N} \left(x_*; \mu^A_{ij}, \boldsymbol\Sigma_{A} \right) - \sum_{i,j} \sum_{k,l} w^{'AA}_{ijkl} \mathcal{N} \left(x_*; \mu^{AA}_{ijkl}, \boldsymbol\Sigma_{AA} \right)
\end{align}

where\\
$\boldsymbol\Sigma_{A} = (2\textbf{W}^{-1} + \boldsymbol\Sigma_{\pi}^{-1})^{-1}$\\
$\mu^{A}_{ij} = \boldsymbol\Sigma_{A} \left( \textbf{W}^{-1} (X_i + X_j) + \boldsymbol\Sigma_{\pi}^{-1} \mu_{\pi} \right)$\\
$w^{'A}_{ij} = w^{'v}_{ij} \mathcal{N} \left(\frac{X_i + X_j}{2}; \mu_{\pi}, \frac{\textbf{W}}{2} + \boldsymbol\Sigma_{\pi} \right)$

$\boldsymbol\Sigma_{AA} = (4W^{-1} + \boldsymbol\Sigma_{\pi}^{-1})^{-1}$\\
$\mu^{AA}_{ijkl} = \boldsymbol\Sigma_{A} \left( \textbf{W}^{-1} (X_i + X_j + X_k + X_l) + \boldsymbol\Sigma_{\pi}^{-1} \mu_{\pi} \right)$\\
$w^{'AA}_{ijkl} = w^{'vv}_{ijkl} \mathcal{N} \left(\frac{X_i + X_j + X_k + X_l}{4}; \mu_{\pi}, \frac{\textbf{W}}{4} + \boldsymbol\Sigma_{\pi} \right)$

\subsubsection{Normalising constant and PDF}
The normalising constant can be obtained via the integral:
\begin{align}
\begin{split}
    Z_A
    &= \int A(x_*) dx_*\\
    &= \sum_{i,j} w^{'A}_{ij} - \sum_{i,j} \sum_{k,l} w^{'AA}_{ijkl}
\end{split}
\end{align}

Thus, the normalised acquisition function as PDF $p_A$ is as follows:
\begin{align}
    p_A(x_*) &= \tilde A(x_*)\\
    &= \frac{C^L_\textbf{y}(x_*) \pi(x_*)}{Z_A}\\
    &= \sum_{i,j} w^A_{ij} \mathcal{N}(x_*; \mu^{A}_{ij}, \boldsymbol\Sigma_{A})
- \sum_{i,j} \sum_{k,l} w^{AA}_{ijkl} \mathcal{N}(x_*; \mu^{AA}_{ijkl}, \boldsymbol\Sigma_{AA})
\end{align}

where
$w^{A}_{ij} = w^{'A}_{ij} / Z_A $\\
$w^{AA}_{ijkl} = w^{'AA}_{ijkl} / Z_A $

\subsubsection{Factorisation trick}

The factorisation trick is set in the conditions where the likelihood is $|\tilde \ell (x)|$,  the distribution of interest $f(x)$ is $|\tilde \ell(x)|\pi(x)$, and the acquiition function is $\tilde C_{\textbf{y}}(x,x)\pi(x)$. We will derive the Gaussian mixture form of this acquisition function.

\begin{align}
  A(x) &= \tilde C_{\textbf{y}}(x,x) \pi(x)\\
  &= \frac{v}{\sqrt{|2\pi \textbf{W}|}}\mathcal{N}(x; \mu_{\pi}, \boldsymbol\Sigma_{\pi}) - v^2 \sum_{ij} \boldsymbol\Omega_{ij} \mathcal{N}(x; \mu^{f}, \boldsymbol\Sigma^{f})
\end{align}

where\\
$\boldsymbol\Sigma^{f} = \left( 2\textbf{W}^{-1} + \boldsymbol\Sigma_{\pi}^{-1} \right)^{-1}$\\
$\boldsymbol\mu^{f} = \boldsymbol\Sigma^{f} \left( \textbf{W}^{-1} (X_i + X_j) + \boldsymbol\Sigma_{\pi}^{-1} \mu_{\pi}\right)$

Then, normalising constant is:
\begin{align}
  Z_A^f = \int A(x) dx = \frac{v}{\sqrt{|2\pi \textbf{W}|}} - v^2 \sum_{ij} \boldsymbol\Omega_{ij}
\end{align}

Therefore, the acquisition function as a probability distribution function $p_A(x)$ is:
\begin{align}
  p_A(x) = \frac{v}{Z_A^f\sqrt{|2\pi \textbf{W}|}}\mathcal{N}(x; \mu_{\pi}, \boldsymbol\Sigma_{\pi}) - \frac{v^2}{Z_A^f} \sum_{ij} \boldsymbol\Omega_{ij} \mathcal{N}(x; \mu^{f}, \boldsymbol\Sigma^{f})\label{eq:factorisation}
\end{align}

\subsection{Efficient sampler}
% this is taken from this: https://stats.stackexchange.com/questions/14481/quantiles-from-the-combination-of-normal-distributions

\subsubsection{Acquisition function as sparse Gaussian mixture sampler}
Eq. \eqref{eq:factorisation} clearly explains the acquisition function can be written as a Gaussian mixture, but it also contains negative components. The first term is obviously positive, and the second term is a mixture of positive and negative components. The condition where the second term becomes positive is $\boldsymbol\Omega_{ij} < 0$. By checking the negativity of the element $\boldsymbol\Omega_{ij}$, we can reduce the number of components by half on average. Then, when we consider sampling from this non-negative acquisition function, the following steps will be performed: First, we sample the index of the component from weighted categorical distribution $\Pi(x)$, and the weights are the one in Eq. \eqref{eq:factorisation}. Then, we sample from the normal distribution that has the same index identified in the first process. These sampling will be repeated until the accumulated number of the sample reaches the same as the recombination sample size $N$. This means the component whose weight is lower than $1/N$ is unlikely to be sampled even once. Therefore, we can dismiss these components with the threshold of $1/N$. Interestingly, the weights of Gaussians vary exponentially. The reduced number of Gaussians is much lower than $n^2$. As such, we can construct the efficient sparse Gaussian mixture sampler of the acquisition function $p^{\prime}_{A}(x)$.

\subsubsection{Sequential Monte Carlo}
Recall from the Eqs (8) - (10) in the main paper, we wish to sample from $g(x) = (1 - r) \pi(x) + r p_A(x)$. We have the efficient sampler $p^{\prime}_{A}(x)$, but $p^{\prime}_{A}(x) \neq  p_{A}(x)$ because $p^{\prime}_{A}(x)$ is the function which is constructed from only positive components of $p_{A}(x)$. Thus, we need to correct this difference via sequential Monte Carlo (SMC). The idea of SMC is simple:

\begin{enumerate}
    \item sample $\textbf{x} \sim p^{\prime}_{A}(x)$, $\textbf{x} \in \mathbb{R}^{rN}$ 
    \item calculate weights $\textbf{w}_\text{smc} = p_{A}(\textbf{x}) / p^{\prime}_{A}(\textbf{x})$
    \item resample from the categorical distribution of the index of $\textbf{x}$ based on $\textbf{w}_\text{smc}$
\end{enumerate}

If $p_{A}(\textbf{x}) \approx p^{\prime}_{A}(\textbf{x})$, the rejected samples in the procedure 3 is minimised. As we formulate $p^{\prime}_{A}(\textbf{x})$ can approximate $p_{A}(\textbf{x})$ well, the number of samples to be rejected is negligibly small. Thus, the number of samples from $p_A(x)$ is slightly smaller than $rN$. The number of samples for $\pi(x)$ in $g(x)$ is adjusted to this fluctuation to keep the partition ratio $r$.

\section{Other BQ modelling}
\subsection{Non-Gaussian Prior}
Non-Gaussian prior distributions can be applied via importance sampling.
\begin{align}
\int \ell(x) \pi(x) 
&= \int \ell(x) \frac{\pi(x)}{g(x)} g(x) dx \\
&= \int \ell^\prime(x) g(x) dx
\end{align}
where $\pi(x)$ is the arbitrary prior distribution of interest, $g(x)$ is the proposal distribution of Gaussian (mixture), $\ell^\prime(x) = \ell(x) \pi(x) / g(x)$ is the modified likelihood.
Then, we set the two independent GPs on each of $\ell(x)$ and $\ell^\prime(x)$. Then, both the model evidence $Z = \int \ell^\prime(x) g(x) dx$, and the posterior $p(x) = \ell(x) \pi(x) / Z$ becomes analytical.

\subsection{Non-Gaussian kernel}
WSABI-BQ methods are limited to the squared exponential kernel in the likelihood modelling. However, other BQ modelling permits the selection of different kernels. For instance, there are the existing works on tractable BQ modelling with kernels of Matérn \citep{briol2019probabilistic}, Wendland \citep{oates2016controlled}, Gegenbauer \citep{briol2019probabilistic}, Trigonometric (Integration by parts), splines \citep{wahba1990spline} polynomial \citep{briol2015frank}, and gradient-based kernel \citep{oates2017control}. See details in \citep{briol2019probabilistic}.

\subsection{RCHQ for Non-Gaussian prior and kernel}
RCHQ permits the integral estimation via non-Gaussian prior and/or kernel without bespoke modelling like the above techniques.
\begin{align}
    \textbf{X}_\text{quad}, \textbf{w}_\text{quad} &= \text{RCHQ}(\text{BQmodel}, \text{sampler})\\
    \mathbb{E}[\ell(x)\pi(x)] &= \textbf{w}_\text{quad} m^L_\textbf{y}(\textbf{X}_\text{quad})\\
    \mathbb{V}\text{ar}[\ell(x)\pi(x)] &= 
    \textbf{w}_\text{rec}^\top  C^L_\textbf{y}(\textbf{x}_\text{rec}, \textbf{x}_\text{rec})\textbf{w}_\text{rec} -
    2\,\textbf{w}_\text{rec}^\top  C^L_\textbf{y}(\textbf{x}_\text{rec}, \textbf{x}_\text{quad}) \textbf{w}_\text{quad}
    + \textbf{w}_\text{quad}^\top  C^L_\textbf{y}(\textbf{x}_\text{quad}, \textbf{x}_\text{quad})\textbf{w}_\text{quad}
\end{align}

\subsection{Vanilla BQ model (VBQ)}
\subsubsection{Expectation}
\begin{align}
\int m_{\ell_0}(x)\pi(x)dx 
&= v \int \mathcal{N}(x; \textbf{X}, \textbf{W}) \mathcal{N}(x; \mu_\pi, \boldsymbol\Sigma_\pi) dx \boldsymbol\omega\\ 
&= v \mathcal{N}(\textbf{X}; \mu_\pi, \textbf{W}+\boldsymbol\Sigma_\pi)\boldsymbol\omega\\ 
\end{align}

\subsubsection{Acquisition function}
\begin{align}
A_\text{unnormalised}(x)
&= C(x,x)\pi(x)\\
&= K(x,x)\pi(x) - K(x,\textbf{X})K(\textbf{X}, \textbf{X})^{-1}K(\textbf{X}, x)\pi(x)\\
&= \mathcal{N}(x; x, \textbf{W})\mathcal{N}(x; \mu_\pi, \boldsymbol\Sigma_\pi) - v^2\mathcal{N}(x; \mu_\pi, \boldsymbol\Sigma_\pi) \mathcal{N}(x;\textbf{X}, \textbf{W})K(\textbf{X}, \textbf{X})^{-1} \mathcal{N}(x;\textbf{X}, \textbf{W})^\top\\
&= \frac{v}{\sqrt{|2\pi\textbf{W}|}}\mathcal{N}(x; \mu_\pi, \boldsymbol\Sigma_\pi) - v^2 \sum_{i,j} \Omega_{ij} \mathcal{N}(\mu_\pi; X_i, \textbf{W} + \boldsymbol\Sigma_\pi) \mathcal{N}(x;X^\prime_i, \textbf{W}^\prime)\mathcal{N}(x;X_j, \textbf{W}) \\
&= \frac{v}{\sqrt{|2\pi\textbf{W}|}}\mathcal{N}(x; \mu_\pi, \boldsymbol\Sigma_\pi) - v^2 \sum_{i,j} \Omega_{ij} \mathcal{N}(\mu_\pi; X_i, \textbf{W} + \boldsymbol\Sigma_\pi) \mathcal{N}(X_j;X^\prime_i, \textbf{W} + \textbf{W}^\prime)\mathcal{N}(x;X^{\prime\prime}_{ij}, \textbf{W}^{\prime\prime})\\
&= \frac{v}{\sqrt{|2\pi\textbf{W}|}}\mathcal{N}(x; \mu_\pi, \boldsymbol\Sigma_\pi) - \sum_{i,j}w_{ij}\mathcal{N}(x;X^{\prime\prime}_{ij}, \textbf{W}^{\prime\prime})\\
\end{align}

where
\begin{align}
\Omega_{ij} &:= K(\textbf{X}, \textbf{X})^{-1}\\
w_i &:= v^2 \Omega_{ij} \mathcal{N}(\mu_\pi; X_i, \textbf{W} + \boldsymbol\Sigma_\pi) \mathcal{N}(X_j;X^\prime_i, \textbf{W} + \textbf{W}^\prime)\\
\textbf{W}^\prime
&= (\textbf{W}^{-1} + \boldsymbol\Sigma_\pi^{-1})^{-1}
\\
X_i^\prime
&= \textbf{W}^\prime(
\textbf{W}^{-1} X_i+ \boldsymbol\Sigma_\pi^{-1} \mu_\pi
)
\\
\textbf{W}^{\prime\prime}
&= \left(
\textbf{W}^{\prime -1} +
\textbf{W}^{-1}
\right)^{-1}
\\
X_{ij}^{\prime\prime}
&= \textbf{W}^{\prime\prime}
\left(
\textbf{W}^{\prime -1} X^\prime_i+ \textbf{W}^{-1} X_j
\right)
\\
\\
\end{align}

Then, the normalised acquisition function $p_A(x)$ as a probability distribution is as follows:
\begin{align}
P_A(x) &:= A_\text{unnormalised}(x) / Z_A
\\
&= \frac{v}{Z_A\sqrt{|2\pi\textbf{W}|}}\mathcal{N}(x; \mu_\pi, \boldsymbol\Sigma_\pi) - \sum_{i,j} \frac{w_i}{Z_A}\mathcal{N}(x;X^{\prime\prime}_{ij}, \textbf{W}^{\prime\prime})\\
\end{align}

where
\begin{align}
Z_A &= \int A_\text{unnormalised}(x) dx\\
&= \frac{v}{\sqrt{|2\pi\textbf{W}|}} \int\mathcal{N}(x; \mu_\pi, \boldsymbol\Sigma_\pi)dx - v^2 \sum_{i,j} \Omega_{ij} \mathcal{N}(\mu_\pi; X_i, \textbf{W} + \boldsymbol\Sigma_\pi)\int \mathcal{N}(x;X^\prime_i, \textbf{W}^\prime)\mathcal{N}(x;X_j, \textbf{W}) dx\\
&= \frac{v}{\sqrt{|2\pi\textbf{W}|}} - v^2  \sum_{i,j} \Omega_{ij} \mathcal{N}(\mu_\pi; X_i, \textbf{W} + \boldsymbol\Sigma_\pi) \mathcal{N}(X_j;X^\prime_i, \textbf{W} + \textbf{W}^\prime)\\
\end{align}

\subsection{Log-GP BQ modelling (BBQ)}
\subsubsection{BBQ modelling}
The doubly-Bayesian quadrature (BBQ) is modelled with log-warped GPs as follows (see details in the paper \citep{osborne2012active}):

\paragraph{Set three GPs}
\begin{align}
p(\ell_0|\textbf{D}) &\sim \mathcal{GP}(\ell_0; m_{\ell_0}(x), C_{\ell_0}(x, x^\prime))\\
p(\log \ell_0|\textbf{D}) &\sim \mathcal{GP}(\log \ell_0; m_{\log \ell_0}(x), C_{\log \ell_0}(x, x^\prime))\\
p(\Delta_{\log \ell_0}|\textbf{D}) &\sim \mathcal{GP}(\Delta_{\log \ell_0}; m_{\Delta}(x), C_{\Delta}(x, x^\prime))\\
\end{align}

\paragraph{Definitions}
\begin{align}
\exp(\log \ell(x)) &\approx \exp(\log \ell_0(x)) + \exp(\log \ell_0(x))(\log \ell(x) - \log \ell_0(x))\\
\ell_0 &:= m_{\ell_0}\\
\Delta_{\log \ell_0} &:= m_{\log \ell_0} - \log \ell_0
=m_{\log \ell_0} - \log (m_{\ell_0})\\
m_\ell &= m_{\ell_0} + m_{\ell_0}m_{\Delta}(x)
\end{align}

\paragraph{Expectation}
\begin{align}
\mathbb{E}[Z|\textbf{D}] &= \int m_{\ell_0}(x)\pi(x)dx + \int m_{\ell_0}(x)m_{\Delta}(x)\pi(x)dx \\
\end{align}

The first term is as follows:
\begin{align}
\int m_{\ell_0}(x)\pi(x)dx
&= v \int \mathcal{N}(x; \textbf{X}, \textbf{W}) \mathcal{N}(x; \mu_\pi, \boldsymbol\Sigma_\pi) dx \boldsymbol\omega\\
&= v \mathcal{N}(\textbf{X}; \mu_\pi, \textbf{W}+\boldsymbol\Sigma_\pi)\boldsymbol\omega\\
\end{align}

where
\begin{align}
\boldsymbol\omega &= K(\textbf{X}, \textbf{X})^{-1}\ell_0(\textbf{X})\\
K(x, \textbf{X}) &= v\mathcal{N}(x; \textbf{X}, \textbf{W})
\end{align}

The second term is as follows:
\begin{align}
&\int m_{\ell_0}(x)\Delta_{\log \ell_0}(x)p(x)dx\\
&= vv^\Delta \boldsymbol\omega^\top  \int \mathcal{N}(x; \textbf{X}, \textbf{W})^\top \mathcal{N}(x; \textbf{X}^\Delta, \textbf{W}^\Delta) \mathcal{N}(x; \mu_\pi, \boldsymbol\Sigma_\pi) dx \boldsymbol\omega^\Delta\\
&= vv^\Delta \boldsymbol\omega^\top  \mathcal{N}(\textbf{X}^\top - \textbf{X}^\Delta, \textbf{0}, \textbf{W} + \textbf{W}^\Delta) \int \mathcal{N}(x;
\boldsymbol\mu^\Delta, \boldsymbol\Sigma^\Delta) \mathcal{N}(x; \mu_\pi, \boldsymbol\Sigma_\pi) dx \boldsymbol\omega^\Delta\\
&= vv^\Delta \boldsymbol\omega^\top  \mathcal{N}(\textbf{X}^\top - \textbf{X}^\Delta, \textbf{0}, \textbf{W} + \textbf{W}^\Delta) \mathcal{N}(\boldsymbol\mu^\Delta; \mu_\pi, \boldsymbol\Sigma_\pi + \boldsymbol\Sigma^\Delta) \boldsymbol\omega^\Delta\\
\end{align}

where
\begin{align}
\boldsymbol\omega^\Delta &= K(\textbf{X}^\Delta, \textbf{X}^\Delta)^{-1}\Delta_{\log \ell_0}(\textbf{X}^\Delta)\\
\mu^\Delta &= [\textbf{W}^{-1} + \textbf{W}^{\Delta, -1}]^{-1}(
\textbf{W}^{-1}\textbf{X} + \textbf{W}^{\Delta, -1}\textbf{X}^\Delta)\\
\boldsymbol\Sigma^\Delta &= [\textbf{W}^{-1} + \textbf{W}^{\Delta, -1}]^{-1}\\
\end{align}

$\textbf{X}^\Delta$ is the observed data for the correlation factor $\Delta_{\log \ell_0}$, which includes not only $\textbf{X}$ but also the additional data points via $m_{\log \ell_0} - \log (m_{\ell_0})$, with GPs calculation.

\subsubsection{Sampling for BBQ}
We apply BASQ-VBQ sampling scheme for log-GP $\text{log}\ell_0$, then calculate the others as post-process. Therefore, the sampling cost is similar to the VBQ, whereas the integral estimation as post-process is more expensive than VBQ.

\section{Experimental details}
\subsection{Synthetic problems}

\subsubsection{Quadrature hyperparameters}
The initial quadrature hyperparameters are as follows:\\
A kernel length scale $l = 2$\\
A kernel variance $v^\prime = 2$\\
Recombination sample size $N = 20,000$ \\
Nystr{\"o}m sample size $M = N/100$ \\
Supersample ratio $r_\text{super} = 100$\\
Proposal distribution $g(x)$ partition ratio $r = 0.5$\\

The supersample ratio $r_\text{super}$ is the ratio of supersamples for SMC sampling of acquisition function against the recombination sample size $N$.

A kernel length scale and a kernel variance are important for selecting the samples in the first batch. Nevertheless, these parameters are updated via type-II MLE optimisation after the second round. Nystr{\"o}m sample size must be larger than the batch size $n$, and the recombination sample size is preferred to satisfy $N \gg M$. Larger $N$ and $M$ give more accurate sample selection via kernel quadrature. However, larger subsamples result in a longer wall-time. We do not need to change the values as long as the integral converged to the designated criterion. When longer computational time is allowed, or likelihood is expensive enough to regard recombination time as negligible, larger $N$, $M$ will give us a faster convergence rate.

The partition ratio $r$ is the only hyperparamter that affects the convergence sensitively. The optimal value depends the integrand and it is challenging to know the optimal value before running. As we derived in Lemma \ref{lem:is}, $\sqrt{C^L_\textbf{y}}\pi(x)$ gives the optimal upper bound. $r=0.5$ is a good approximation of this optimal proposal distribution: $g(x) = (1-r) \pi(x) +r C^L_\textbf{y}\pi(x) = \left\{ (1-r) + rC^L_\textbf{y} \right\} \pi(x)$. Here, the linearisation gives the approximation $\sqrt{C^L_\textbf{y}} = \sqrt{1 + (C^L_\textbf{y} -1)} \approx 1 + \frac{C^L_\textbf{y} -1}{2} = 0.5 + 0.5C^L_\textbf{y}$. Therefore, $(0.5 + 0.5C^L_\textbf{y})\pi(x) \approx \sqrt{C^L_\textbf{y}}\pi(x)$. Thus, $r=0.5$ is a safe choice.

\subsubsection{Gaussian mixture}
The likelihood function of the Gaussian mixture used in Figure 1 in the main paper is expressed as:
\begin{align}
    \ell_{\text{true}}(x) &= \sum_{i=1}^n w_i \mathcal{N}(x; \mu_i, \boldsymbol\Sigma_i)\\
    w_i &= \mathcal{N}(\mu_i; \mu_{\pi}, \boldsymbol\Sigma_i + \boldsymbol\Sigma_{\pi})^{-1}\\
    Z_{\text{true}} &= \int \ell_{\text{true}}(x) \pi(x) dx,\\
    &= \sum_{i=1}^n w_i \int \mathcal{N}(x; \mu_i, \boldsymbol\Sigma_i) \mathcal{N}(x; \mu_{\pi}, \boldsymbol\Sigma_{\pi}) dx\\
    &= \sum_{i=1}^n w_i \mathcal{N}(\mu_i; \mu_{\pi}, \boldsymbol\Sigma_i + \boldsymbol\Sigma_{\pi})\\
    &= 1
\end{align}

where\\
$\mu_\pi = \textbf{0}$\\
$\boldsymbol\Sigma_\pi = 2\textbf{I}$\\
$\pi(x) = \mathcal{N}(x; \mu_{\pi}, \boldsymbol\Sigma_{\pi})$\\

The prior is the same throughout the synthetic problems.

\subsubsection{Branin-Hoo function}
The Branin-Hoo function in Figure 2 in the main paper is expressed as:
\begin{align}
    \ell_{\text{true}}(x) &= \prod_{i=1}^2 \frac{\left[ \sin(x_i) + \frac{1}{2}\cos(3x_i)\right]^2}{(\frac{1}{2}x_i)^2+\frac{3}{10}},
    \quad x \in \mathbb{R}^2\\
    Z_{\text{true}} &= \int \ell_{\text{true}}(x) \pi(x) dx\\
    &= 0.955728^2\\
    &\approx 0.913416
\end{align}

\subsubsection{Ackley function}
The Ackley function in Figure 2 in the main paper is expressed as:
\begin{align}
    \ell_{\text{true}}(x) &= -20\exp \left(-\frac{1}{5}\sqrt{\frac{1}{2} \sum_{i=1}^2x_i^2} \right) + 
    \exp \left(\frac{1}{2} \sum_{i=1}^2 \cos(2 \pi x_i) \right) + 20, \quad x \in \mathbb{R}^2\\
    Z_{\text{true}} &= \int \ell_{\text{true}}(x) \pi(x) dx\\
    &\approx 5.43478
\end{align}

\subsubsection{Oscillatory function}
The Oscillatory function in Figure 2 in the main paper is expressed as:
\begin{align}
    \ell_{\text{true}}(x) &= \cos \left(2 \pi + 5 \sum_{i=1}^2 x_i \right)+1, \quad x \in \mathbb{R}^2\\
    Z_{\text{true}} &= \int \ell_{\text{true}}(x) \pi(x) dx\\
    &= 1
\end{align}

\subsubsection{Additional experiments}
\paragraph{Dimensional study in Gaussian mixture likelihood}
\begin{figure}
  \centering
  \includegraphics[width=1\textwidth,center]{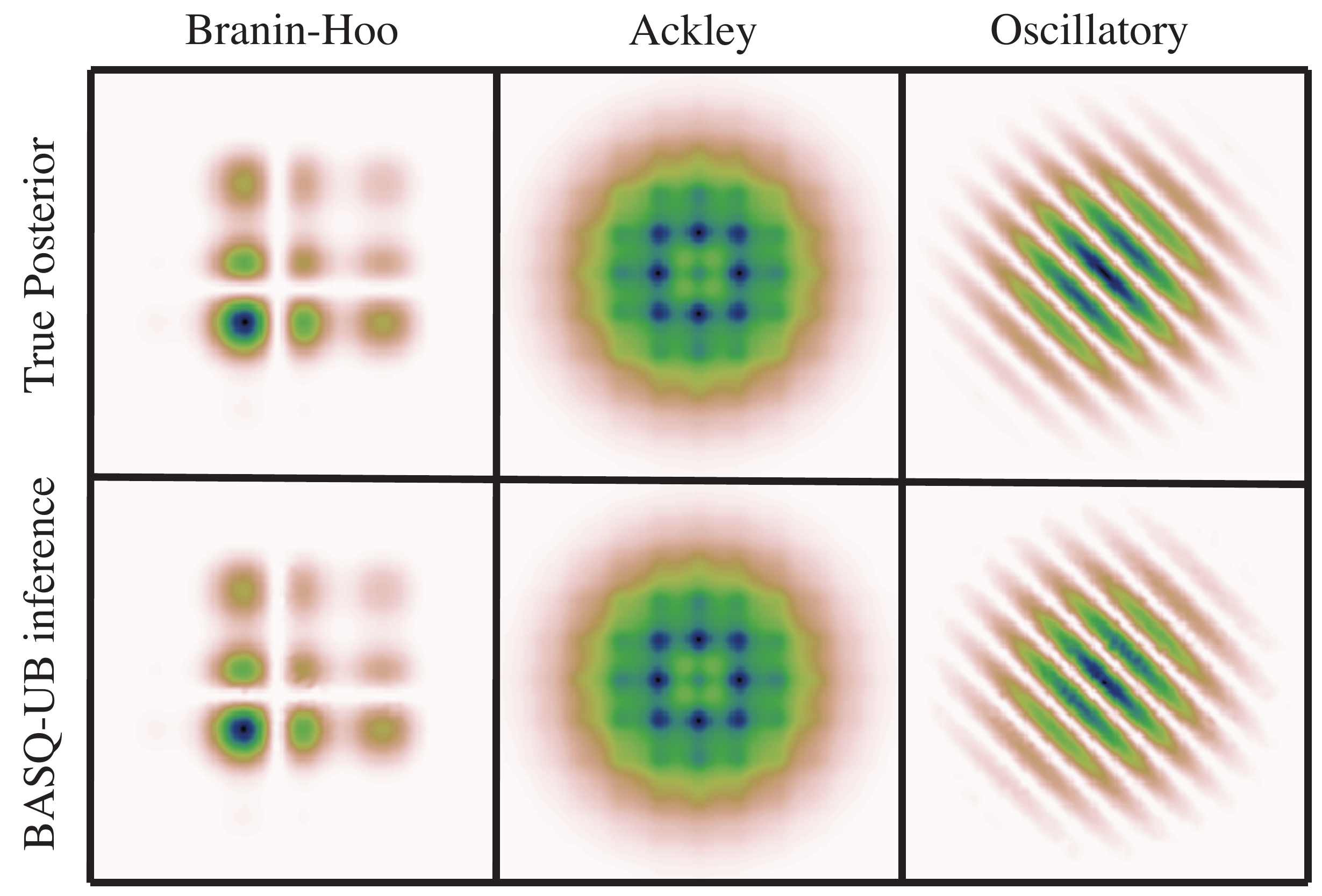}
  \caption{Performance comparison with N-dimensional Gaussian mixture likelihood function. (a) dimension study, (b) convergence rate, and (c) wall time vs MAE of integral. (a) varies from 2 to 16 dimensions, (b) and (c) are 10 dimensional Gaussian mixture.} \label{fig:dimension}
\end{figure}

Figure ~\ref{fig:dimension}(a) shows the dimension study of Gaussian mixture likelihood. The BASQ and BQ are conditioned at the same time budget (200 seconds). The higher dimension gives a more inaccurate estimation. From this result, we recommend using BASQ with fewer than 16 dimensions.

\paragraph{Ablation study}
We investigated the influence of the approximation we adopted using 10 dimensional Gaussian mixture likelihood. The compared models are as follows:
\begin{enumerate}
    \item Exact sampler (without factorisation trick)
    \item Provable recombination (without LP solver)
\end{enumerate}

The exact sampler without the factorisation trick is the one that exactly follows the Eqs. (8) - (10) of the main paper. That is, the distribution of interest $f(x)$ is the prior $\pi(x)$. In addition, the kernel for the acquisition function is an unwarped $C^L_\textbf{y}$, which is computationally expensive. Next, the provable recombination algorithm is the one introduced in \citep{tch15} with the best known computational complexity. As explained in the Background section of the main paper, our BASQ implementation is based on an LP solver (Gurobi [31] for this time) with empirically faster computational time. We compared the influence of these approximations.

Figure ~\ref{fig:dimension}(b) illustrates that these approximations are not affecting the convergence rate in the sample efficiency. However, when compared to the wall-clock time (Figure ~\ref{fig:dimension}(c)), the exact sampler without the factorisation trick is apparently slow to converge. Moreover, the provable recombination algorithm is slower than an LP solver implementation. Thus, the number of samples the provable recombination algorithm per wall time is much smaller than the LP solver. Therefore, our BASQ standard solver delivers solid empirical performance.

\paragraph{Qualitative evaluation of posterior inference}
\begin{figure}
  \centering
  \includegraphics[width=0.6\textwidth,center]{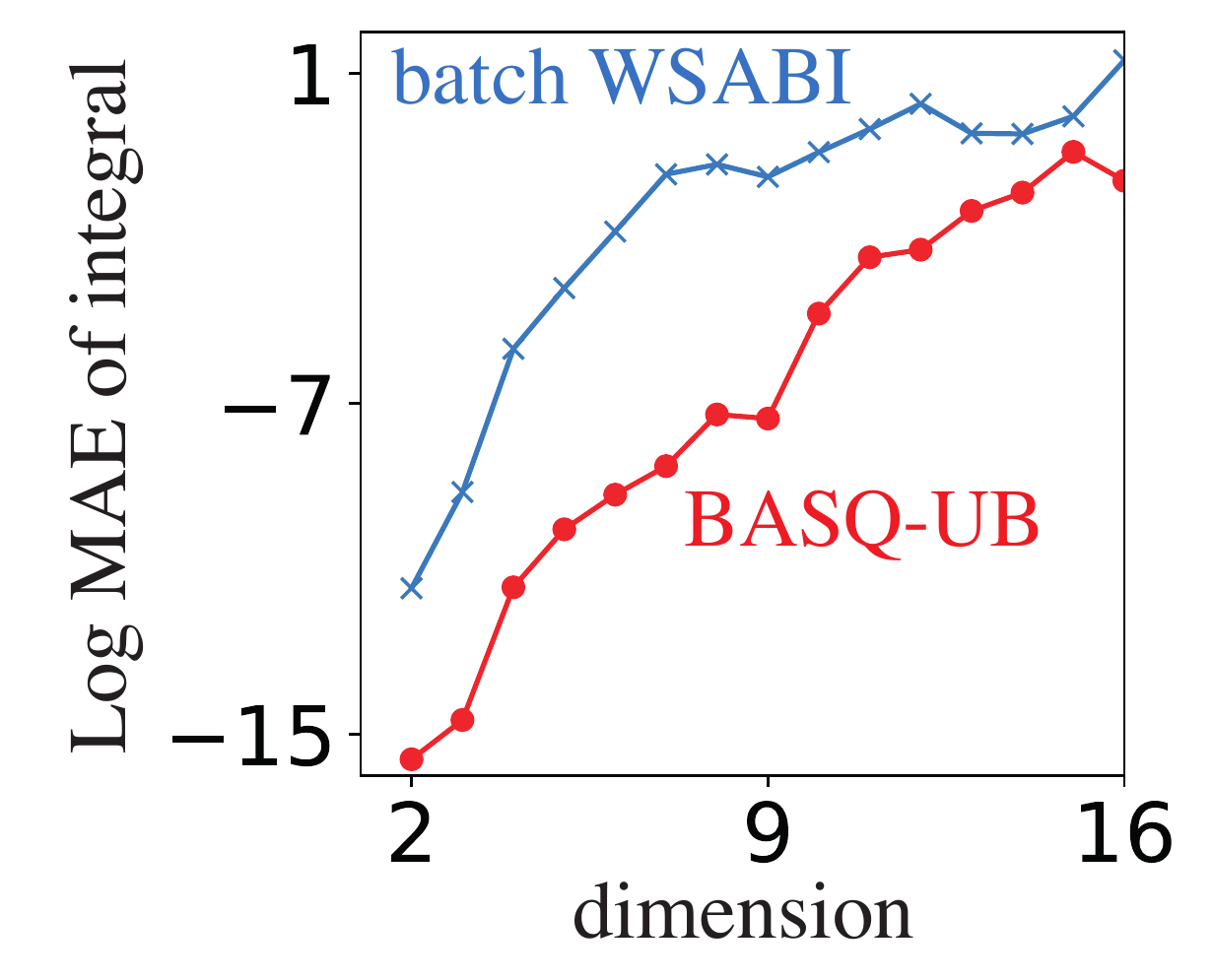}
  \caption{Qualitative evaluation of posterior inference in synthetic problems} \label{fig:synthetic}
\end{figure}

Figure ~\ref{fig:synthetic} shows the qualitative evaluation of joint posterior inference after 200 seconds passed against the analytical true posterior. The estimated posterior shape is exactly the same as the ground truth.

\subsection{Real-world problems}
\subsubsection{Battery simulator}
\paragraph{Background}
Single Particle Model with electrolyte dynamics (SPMe) is a commonly-used lithium-ion battery simulator to predict the voltage response at given excitation current time-series data. Estimating SPMe parameters from observations are well known for ill-conditioned problem because this model is overparameterised \citep{bizeray2019identifiability}. In the physical model, we need to separate the anode and cathode internal states to represent actual cell components. However, when it comes to predicting the voltage response, this separation into two components is redundant. Except for extreme conditions such as low temperature, most voltage responses can be expressed with a single component. Therefore, the parameters of cathode and anode often have a perfect negative correlation, meaning an arbitrary combination of cathode and anode parameters can reconstruct the exactly same voltage profile. As such, point estimation means nothing in these cases. Bayesian inference can capture this negative correlation as covariance. Therefore, Bayesian inference is a natural choice for parameter estimation in the battery simulator. Moreover, there are many plausible battery simulators with differing levels of approximation. Selecting the model satisfying both predictability and a minimal number of parameters is crucial for faster calculation, particularly in setting up the control simulator. Therefore, Bayesian model selection with model evidence is essential. The experimental setup is basically following \citep{aitio2020bayesian}.

\paragraph{Problem setting}
We wish to infer the posterior distribution of 3 simulation parameters $(D_n, D_p, \sigma_n)$, where $D_n$ is the diffusivity on anode, $D_p$ is the diffusivity on cathode, $\sigma_n$ is the noise variance of the observed data. We have the observed time-series voltage $\textbf{y}$ and exciation profiles $\textbf{i}$ as training dataset.

The parameter inference is modelled as follows:
\begin{align}
  y_* &= \text{Sim}(x_*, i_*)\\
  \pi(x_*) &= \text{Lognormal}(x_*; \mu_{\pi}, \boldsymbol\Sigma_{\pi})\\
  \ell_{\text{true}}(x_*) &= \mathcal{N} \left[ \text{Sim}(x_*, \textbf{i}); \textbf{y}, \sigma_n \textbf{I} \right]
\end{align}

where\\
$\mu_\pi = [1.38,0, -20.25]$\\
$\boldsymbol\Sigma_\pi = \text{diag}([0.03,0.001, 0.001])$\\
in the logarithmic space.

\paragraph{Parameters}
The observed data $\textbf{y}$ and $\textbf{i}$ are generated by the simulator with multiharmonic sinusoidal excitation current defined as:

\begin{align}
    \textbf{i} &= 0.132671 \left[ \sin(1/5 \pi t) + \sin(2 \pi t) + \sin(20 \pi t) + \sin(200 \pi t) \right]\\
    \textbf{y} &= \text{Sim}(x_\text{true}, \textbf{i}) + \sqrt{\sigma_n} \mathcal{U}[0,1]
\end{align}

where\\
$t$ is discretised for 10 seconds with the sampling rate of 0.00025 seconds, resulting in 40,000 data points.\\
$x_\text{true}$ = [
$\exp(1.361) \times 10^{-14}$,
$\exp(0) \times 10^{-13}$,
$\exp(-20.25) \times 10^{-10}$
]

\paragraph{Metric}
The posterior distribution is evaluated via RMSE between true and inferred conditional posterior on each parameter. The RMSE is calculated on 50 grid samples for each dimension so as to slice the maximum value of the joint posterior. Each 50 grid samples are equally-spaced and bounded with the following boundaries:\\
bounds $= [1.1,1.7], [-0.075, 0.08], [-20.3,-20.2]$\\
where the boundaries are given by [lower, upper] in the logarithmic space.

\subsubsection{Phase-field model}
\paragraph{Background}
The PFM is a flexible time-evolving interfacial physical model that can easily incorporate the multi-physical energy \citep{kim1999phase}. In this dataset, the PFM is applied to the simulation of spinodal decomposition, which is the self-organised nanostructure in the bistable Fe-Cr alloy at high temperatures. Spinodal decomposition is an inherently stochastic process, making characterisation challenging \citep{matsuura2021adjoint}. Therefore, Bayesian model selection is promising for estimating its parameter and determining the model physics component.

\paragraph{Problem setting}
We wish to infer the posterior distribution of 4 simulation parameters $(T, L_{cT}, n_B, L_g)$, 
where $T$ is the temperature, $L_{cT}$ is the interaction parameter that defines the interaction between composition and temperature, $n_B$ is the number of Bohr magnetons per atom, and $L_g$ is the gradient energy coefficient.
We have the observed time-series 2-dimensional images $\textbf{y}$.

The parameter inference is modelled as follows:
\begin{align}
  y_* &= \text{Sim}(x_*)\\
  \pi(x_*) &= \text{Lognormal}(x_*; \mu_{\pi}, \boldsymbol\Sigma_{\pi})\\
  \ell_{\text{true}}(x_*) &= \mathcal{N} \left[ \text{Sim}(x_*); \textbf{y}, \sigma_n \textbf{I} \right]
\end{align}

where\\
$\sigma_n = 10^{-4}$\\
$\mu_\pi = [1.91, 0.718, 0.798, 0.693]$\\
$\boldsymbol\Sigma_\pi = \text{diag}([0.0003, 0.00006, 0.0001, 0.0001])$\\
in the logarithmic space.

\paragraph{Parameters}
The observed data $\textbf{y}$ is generated by the simulator defined as:

\begin{align}
    \textbf{y} &= \text{Sim}(x_\text{true}) + \sqrt{\sigma_n} \mathcal{U}[0,1]
\end{align}

where\\
$\textbf{y}$ is discretised in both spatially and time-domain. Time domain is discretised for 5000 seconds with the sampling rate of 1 seconds, resulting in 5,000 data points. 2-dimensional space is discretised for $64 \times 64$ $\text{nm}^2$, with  $64 \times 64$ $\text{nm}^2$ pixels. The total data points are $64 \times 64 \times 5,000 = 20,480,000$.\\
$x_\text{true}$ = [
$\exp(1.90657514) \times 10^{2}$,
$\exp(0.71783979) \times 10^{4}$,
$\exp(0.7975072)$,
$\exp(0.69314718) \times 10^{-15}$
]

\paragraph{Metric}
The posterior distribution is evaluated via RMSE between true and inferred conditional posterior on each parameter. The RMSE is calculated on 50 grid samples for each dimension so as to slice the maximum value of the joint posterior. Each 50 grid samples are equally-spaced and bounded with the following boundaries:\\
bounds $= [1.87,1.94],[0.69, 0.73],[0.77, 0.83],[0.68, 0.73]$\\
where the boundaries are given by [lower, upper] in the logarithmic space.

\subsubsection{Hyperparameter marginalisation of hierarchical GP}
\paragraph{Background}
The hierarchical GP model was designed for analysing the large-scale battery time-series dataset from solar off-grid system field data all over the African continent \citep{aitio2021predicting}. The field data contains the information of time-series operating conditions $(I, T, V)$, where $I$ is the excitation current, $T$ is the temperature, and $V$ is the voltage. We wish to estimate the state of health (SoH) from these field data, achieving the preventive battery replacement before it fails for the convenience of those who rely on the power system for their living. However, estimating the state of health is challenging because the raw data $(I, T, V)$ is not correlated to the battery health. There are several definitions of SoH, but the internal resistance of a battery $R$ is adopted in \citep{aitio2021predicting}. In the usual circuit element, resistance can be easily calculated from $R = V/I$ via Ohm's law. However, the battery internal resistance $R$ is way more complex. Battery internal resistance $R$ is a function of $(t, I, T, c)$, where $t$ is time, $c$ is the acid concentration. Furthermore, there are two factors of resistance variation; ionic polarisation and aging. To incorporate these physical insights to the machine learning model, \citep{aitio2021predicting} is adopted the hierarchical GP model. First, they adopted the additive kernel of a squared exponential kernel and a Wiener velocity kernel to divide the ionic polarisation effect and aging effect. Second, they adopted the hierarchical GPs to model $V$ to divide into $R$-dependent GP and non-$R$-dependent GP to incorporate the Open Circuit Voltage-State of Charge (OCV-SOC) relationship.

\paragraph{Problem setting}
We wish to infer the hyperposterior distribution of 5 GP hyperparameters $(l_T, l_I, l_c, \sigma_0, \sigma_1)$, where $l_T, l_I, l_c$ are the a squared exponential kernel lengthscale of temperature $T$, current $I$, and acid concentration $c$, respectively, and $\sigma_0, \sigma_1$ are the kernel variances of a squared exponential kernel and a Wiener velocity kernel, respectively.
We have the observed time-series dataset of $(I, T, V)$ as $\textbf{y}$.

The hyperposterior inference is based on the energy function $\Phi(x)$ (The details can be found in \citep{aitio2021predicting}, Equation (15) in the Appendix information).

\begin{align}
    \Phi{x} &= -\log p(\textbf{y}|x) - \log p(x)\\
    &=  - \log p(x) + \frac{1}{2}\sum_t \log |S_t(x)| +
    \frac{1}{2}\sum_t e_t^\text{T} S_t(x)^{-1} e_t + 
    \sum_t \frac{n_t}{2} \log 2 \pi
\end{align}

where\\
$p(x) = \text{Lognormal}(x_*; \mu_{\pi}, \boldsymbol\Sigma_{\pi})$ is a hyperprior.\\
$e_t$ is the error vector for each charging segment.\\
$n_t$ is the number of observations in the charging segment.\\
$S_t(x)$ is the innovation covariance for the segment.\\
$\mu_\pi = [3.96, 1.94, 2.79, 2.26, 0.34]$\\
$\boldsymbol\Sigma_\pi = \text{diag}([1,1,1,1,1])$\\
in the logarithmic space.

\paragraph{Metric}
The posterior distribution is evaluated via RMSE between true and inferred conditional posterior on each parameter. The RMSE is calculated on 50 grid samples for each dimension so as to slice the maximum value of the joint posterior. Each 50 grid samples are equally-spaced and bounded with the following boundaries:\\
bounds $= [-10,10],[-10,10],[-10,10],[-10,10],[-10,10]$\\
where the boundaries are given by [lower, upper] in the logarithmic space.

\section{Technical details; Q \& A}
\paragraph{Q1: How does BASQ enable RCHQ to perform the batch selection?}
A1: The trick that achieves parallelisation is the alternately subsampling in section 4.1, not RCHQ itself. While BQ aims to calculate the target integral $Z = \int \ell_\text{true}(x)\pi(x)dx$, RCHQ over a single iteration aims to calculate the empirical integral $Z = \int \ell(x)\pi_\text{emp}(x)dx$ over empirical measure defined by $N$ subsamples $\textbf{X}_\text{rec}$. At each iteration, we greedily select the batch candidates via RCHQ that can minimise the integral variance over the current empirical measure. As we gather more observation data points and update the kernel (GP), the above two integrals approach the same. In other words, any KQ method, including kernel herding, can be applied to the batch selection via this alternately subsampling scheme. Secondly, such a dual quadrature scheme tends to be computationally demanding, but tractable computation and superb sample efficiency of RCHQ permit scalable batch parallelisation.

\paragraph{Q2: Why does RCHQ outperform the kernel herding?}
A2: The reason why RCHQ converges faster than herding is that RCHQ exploits more information than herding. While herding greedily optimises sequentially, RCHQ explicitly exploits the information of the spectral decay of the kernel and the probability distribution, both of which herding neglects. Exploiting the spectral decay corresponds to capturing the approximately finite dimensionality of the kernel. RCHQ adopts the Nyström method for its approximation. This convergence rate superiority can be confirmed in  figure 2(a) in \citep{hayakawa2021positively}. ("N. + emp + opt" refers to RCHQ.) While RCHQ exponentially decays, herding does not show such fast decay in the Gaussian setting. Therefore, BQ with RCHQ can converge faster than BQ with kernel herding, allowing scalable and sample-efficient batch selection.

\paragraph{Q3: Are there some potential areas, if any, where the proposed method performs worse than existing ones?}
A3: Probably yes, there is. The advantage of herding over RCHQ is the computation cost. In the small batch size setting, the difference in the level of convergence between herding and RCHQ is much smaller than in the large batch size $n$. Therefore, herding might perform better than RCHQ in the small batch with a very cheap likelihood case as herding might earn more samples than RCHQ. The comparison against other KQ methods is summarised in table 1 in \citep{hayakawa2021positively}. RCHQ gives a small theoretical bound of the worst-case error with tractable computation cost compared to herding, DPP, CVS, and vanilla BQ.

\paragraph{Q4: What are the pros and cons of RCHQ over the Determinantal Point Process (DPP)?}
DPP considers the correlation correctly, whereas RCHQ assumes i.i.d. However, DPP requires prohibitive computation. Table 1 in \citep{hayakawa2021positively} compares DPP-based KQ \citep{belhadji2019kernel} and RCHQ ("N. + empirical" refers to RCHQ), which clearly shows that RCHQ provides not only tractable computation but also competitive theoretical bound of worst-case error with mathematical proof.

\paragraph{Q5: Why Nyström? Other low-rank approximation possibilities?}
A5: Because Nyström is advantageous to derive convergence based on spectral decay asymptotically and theoretically. The only requirement for the RCHQ is a finite sequence of good test functions, so finite dimensional approximations such as random Fourier features can also be used.

\paragraph{Q6: Theorem 1 does not apply to WSABI-transformed BASQ but to a variant which uses vanilla BQ. Is that correct?}
A6: Yes. Theorem 1 is under the assumptions which the BQ is modelled with vanilla BQ, without batch and hyperparameter updates. However, if we accept the linearisation of WSABI-L and assume that the $\ell(x)$ is (approximately) in the GP over the current iteration, the theoretical analysis is correct.

\paragraph{Q7: Why is the coefficient 0.8 used in the definition of alpha?}

A7: We inherit the coefficient of 0.8 from the original paper \citep{gunter2014sampling}. However, they said the performance is insensitive to the choice of this coefficient, so it is not limited to 0.8 in general.

\paragraph{Q8: Can we apply WSABI for negative integral?}

A8: Yes. $\alpha$ can take negative value, so WSABI transformation can be applied to negative integral case when we apply WSABI to general function integration.

\subsection{Detailed description of the difference between RCHQ and batch WSABI}
\begin{figure}
  \centering
  \includegraphics[width=0.7\textwidth,center]{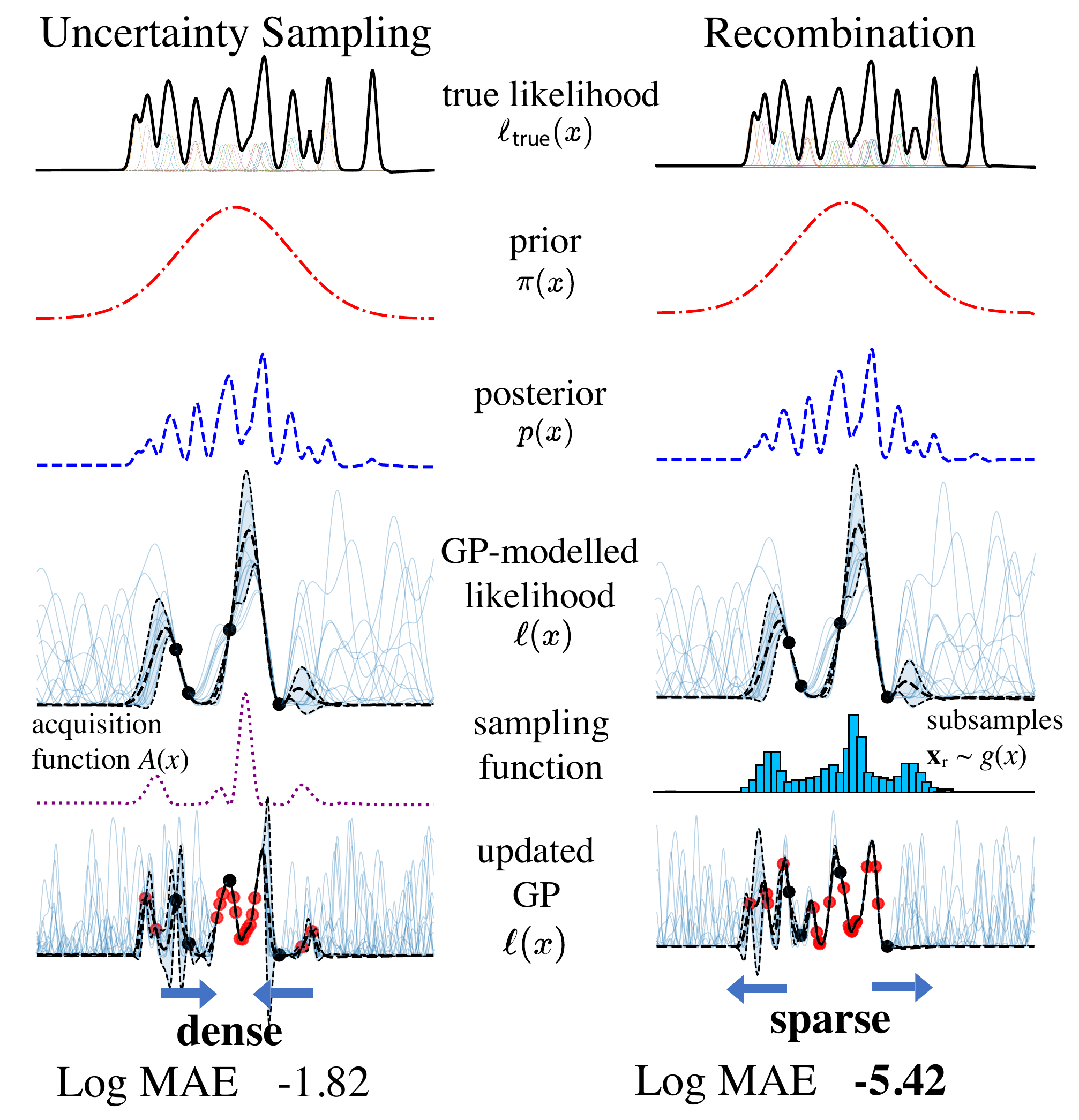}
  \caption{Performance comparison with 1-dimensional Gaussian mixture likelihood function.} \label{fig:details}
\end{figure}

Figure ~\ref{fig:details} shows the performance comparison between the kernel recombination (RCHQ) and the uncertainty sampling (batch WSABI). Firstly, the true likelihood $\ell_\text{true}$ was modelled with the mixture of one-dimensional Gaussians, which was generated under the procedure described in footnote 4 in the main paper. Prior $\pi(x)$ is a broader one-dimensional Gaussian, and the posterior $p(x)$ is also the mixture of Gaussians, thanks to the Gaussianity. While we can access the information of the prior distribution function $\pi(x)$, we cannot access the ones of the posterior $p(x)$ or true likelihood function $\ell_\text{true}$. However, we can query the true likelihood value at a given location $\ell_\text{true}(\textbf{X}_\text{quad})$ with a large batch ($n$ = 16 in this case). Now, we have four observations $n=4$ with black dots. We have constructed the WSABI-L GPs with the given four observations $(\textbf{X}, \textbf{y})$. The mean dotted line represents the mean of posterior predictive distributions $m^L_\textbf{y}(x)$, the blue shaded area shows the mean $\pm$ corrected variance $C^L_\textbf{y}(x,x) \pi(x)$. A myriad of blue lines represents the functions sampled from GP $\ell \sim \mathcal{GP}(\ell; m^L_\textbf{y}(x), C^L_\textbf{y}(x,x))$. The above problem setting is shared with both algorithms.

On the one hand, batch WSABI adopts local penalisation with multi-start optimisation. The acquisition function $A(x)$ for batch WSABI is the uncertainty sampling $\mathbb{V}\text{ar}[\ell(x)\pi(x)] = \pi(x)^2 C_\textbf{y}(x) m_\textbf{y}(x)^2]$ as shown in purple dotted line. We can see four peaks corresponding to the positions of larger variance in WSABI-L GP. Multi-start optimisation generates 100 random samples from prior as multi-starting points, then run a gradient-based optimisation algorithm (L-BFGS) to find the maxima. Then, we take the largest point amongst the solutions. After taking the largest point, we locally penalise the point with Lipschitz Cone. Then the largest peak is split into two peaks with smaller heights. Then, the multi-start optimisation will be applied again to find the next maxima. We will iterate this greedy selection for $n=16$. The selected points are depicted with red dots. The WSABI-L GPs are updated with the given 16 observations. As we can see, the selected 16 batch candidates are concentrated around the largest peak in the acquisition function, resulting in a higher integration error. This is mainly because the local penalisation tends to aggregate around the large peak, where the newly generated penalised peaks still have significant heights. When compared the acquisition function $A(x)$ with the true posterior $p(x)$, we can find that the peak positions between $A(x)$ and $p(x)$ are not so correlated. In other words, local penalisation trusts the acquisition function too much, although the early-stage acquisition function $A(x)$ is not such a reliable information source. Moreover, the multi-start optimisation requires the optimisation loop per the number of seeds, which is computationally demanding. In addition, the possibility of finding the global optima becomes exponentially lower when the dimension is scaled up. Thus, the multi-start optimisation requires exponentially increasing the number of random seeds, although there are still no guarantees to find the global maxima of acquisition function $A(x)$. Therefore, batch WSABI is slow and inefficient in selecting the batch candidates.

On the other hand, RCHQ using the linearised IVR proposal distribution $g(x) = (1-r)\pi(x) + rA(x)$. This is mixed with the prior and GP variance, so it is less dependent on the early-stage $A(x)$. The subsampled histogram depicted with blue bars has similar peaks with $A(x)$, but still, there is room for the possibility of selecting other regions. Then, RCHQ constructs the empirical measure based on these $N$ subsamples and resamples $M$ samples for the Nyström method to construct the finite test functions. The test functions are applied to construct the metric to evaluate the worst-case error. Then, RCHQ selects $n$ batch candidates to minimise the worst-case error over the empirical measure with the kernel recombination. As we can see, the selected 16 batch candidates are sparser and well-captured the true likelihood peaks than local penalisation, resulting in a smaller integration error. Such subsampling is done faster than multi-start optimisation thanks to the efficient sampler, and recombination is also tractable with single LP solver iteration. As such, the RCHQ can select sparser candidates than local penalisation within more tractable computation time.

\end{document}